\documentclass{article}

\PassOptionsToPackage{numbers,sort&compress}{natbib}
\usepackage[preprint]{arxiv_2024}

\usepackage[utf8]{inputenc} 
\usepackage[T1]{fontenc}    
\usepackage{hyperref}       
\usepackage{url}            
\usepackage{booktabs}       
\usepackage{amsfonts}       
\usepackage{nicefrac}       
\usepackage{microtype}      
\usepackage{graphicx}
\usepackage{caption}
\usepackage{amsmath}
\usepackage{amssymb}
\usepackage{mathtools}
\usepackage{wrapfig}

\usepackage[dvipsnames,svgnames,x11names]{xcolor}
\usepackage[utf8]{inputenc} 
\usepackage[T1]{fontenc}    
\usepackage{hyperref}       
\usepackage{url}            
\usepackage{booktabs}       
\usepackage{amsfonts}       
\usepackage{nicefrac}       
\usepackage{microtype}      
\usepackage{multicol}
\usepackage{multirow}
\usepackage{adjustbox}
\usepackage{wrapfig}
\usepackage{changes}
\usepackage{xspace}
\usepackage{float}
\usepackage{cleveref}

\definechangesauthor[color=BrickRed]{SS}

\newenvironment{packed_itemize}
{\begin{itemize}
    \setlength{\itemsep}{1pt}
    \setlength{\parskip}{0pt}
    \setlength{\parsep}{0pt}
}{\end{itemize}}

\newcommand{\filluptopage}[1]{%
  \clearpage
  \loop\ifnum\value{page}<#1\relax
    \null\clearpage
  \repeat
  \loop\ifnum\value{page}=#1\relax
    \null\clearpage
  \repeat
}

\makeatletter
\def\blfootnote{\xdef\@thefnmark{}\@footnotetext}
\makeatother

\newcommand{\shortname}{GenHeld\xspace}
\newcommand{\shortnamethree}{GenHeld3D\xspace}
\newcommand{\shortnametwo}{GenHeld2D\xspace}

\title{\shortname: Generating and Editing Handheld Objects}
\author{
Chaerin Min$^1\quad\quad\quad\quad\quad\quad$
Srinath Sridhar$^1$\thanks{Corresponding Author}\\
$^1$Brown University\\
\texttt{$^1$\{chaerin\_min, srinath\_sridhar\}@brown.edu}\\
\url{https://ivl.cs.brown.edu/research/genheld.html}
}

\begin{document}

\onecolumn
{
\maketitle
\vspace{-0.22in}
{\includegraphics[width=1\linewidth]{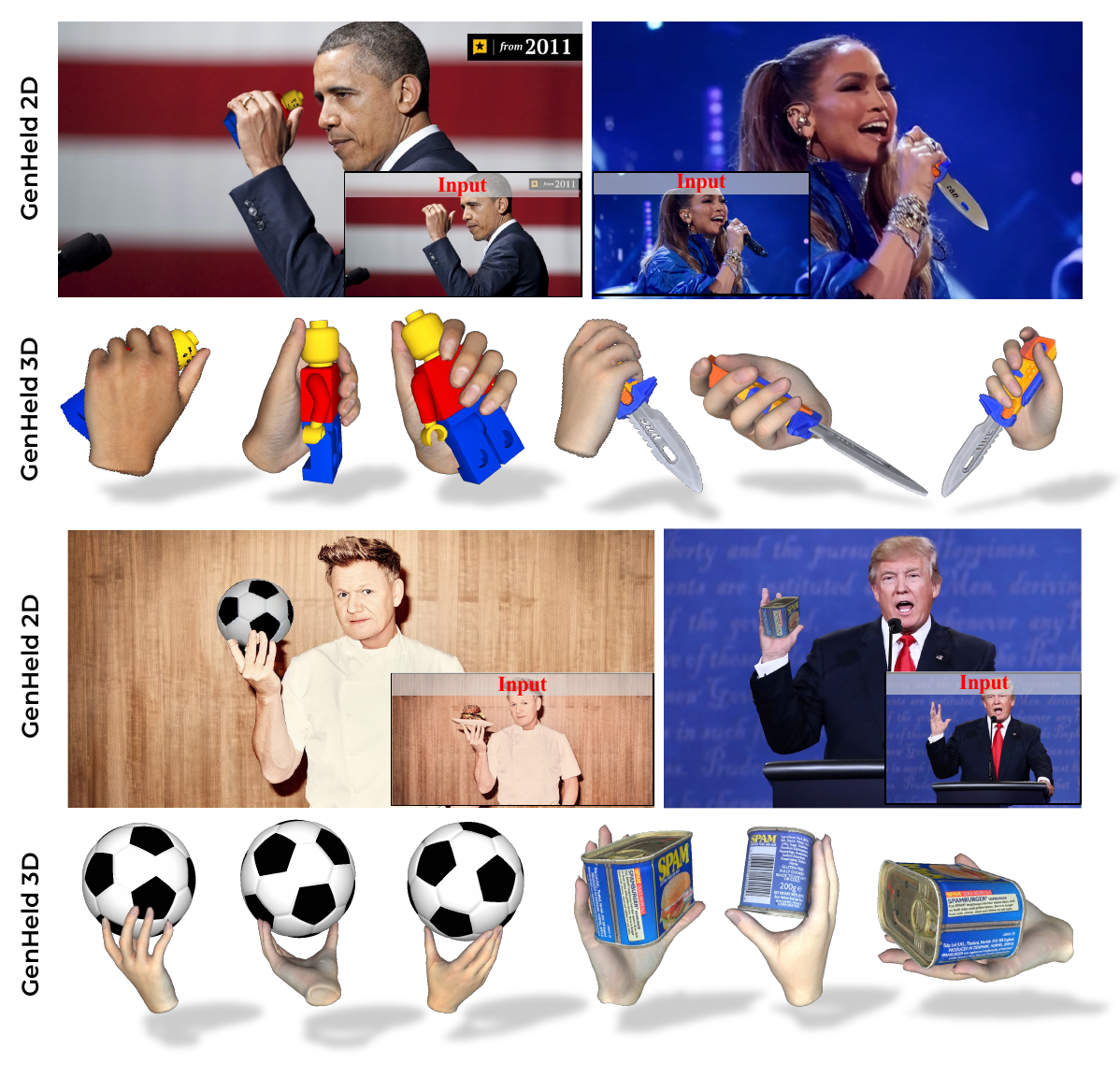}
\captionof{figure}{
We present \shortname, a model to synthesize held objects given 3D hand model or 2D hand image.
\shortnamethree can select plausible and diverse objects from a large object repository~\cite{objaverse}, while \shortnametwo can add or replace existing held objects in images.}
\label{fig:teaser}
}

\begin{abstract}
%
%
Grasping is an important human activity that has long been studied in robotics, computer vision, and cognitive science.
Most existing works study grasping from the perspective of synthesizing hand poses conditioned on 3D or 2D object representations.
We propose \shortname to address the inverse problem of synthesizing held objects conditioned on 3D hand model or 2D image.
Given a 3D model of hand, \shortnamethree can select a plausible held object from a large dataset using compact object representations called \emph{object codes}.
The selected object is then positioned and oriented to form a plausible grasp without changing hand pose.
If only a 2D hand image is available, \shortnametwo can edit this image to add or replace a held object.
\shortnametwo operates by combining the abilities of \shortnamethree with diffusion-based image editing.
Results and experiments show that we outperform baselines and can generate plausible held objects in both 2D and 3D.
%
Our experiments demonstrate that our method achieves high quality and plausibility of held object synthesis in both 3D and 2D.
\end{abstract}
\vspace{-0.5cm}
\section{Introduction}
\label{sec:intro}
Touching, grasping, and manipulating objects constitute a majority of human interactive activities.
Each day, we touch an average of 821 surfaces~\cite{we_touch_821_surfaces} and 140 different objects~\cite{we_touch_140_objects}, enabling us to shape our environments to achieve our physical goals.
Understanding how humans grasp objects can have profound impact in areas like robotics and mixed reality.
Unsurprisingly, this important skill has long been a topic of study in robotics~\cite{classical_grasping, gendexgrasp, se3_diffusionfields, dexdiffuser, realdex, forceclosure, dexgraspnet, crosshand}, computer vision~\cite{text2grasp, semgrasp, hoidiffusion, handiffuser, ho3d, affordancediffusion}, and cognitive science~\cite{grasp_aperture, grip_aperture}.

The focus of most existing work that studies human grasps has been on synthesizing plausible or stable grasping hand poses conditioned on 3D object models~\cite{classical_grasping2, classical_grasping3, classical_grasping4, grasp_survey3,grasptta, grasping_fields, toch, goal, dgrasp, graspd, contactgen, contactedit, cams, graspxl, realdex,hoidiffusion},
or 2D hand images~\cite{affordancediffusion, image_based1, image_based2, image_based3}.
In this problem setting, the object is assumed to be known, and the goal is to generate hand pose and/or shape parameters~\cite{mano} that result in a plausible grasp.
Some methods focus on synthesizing \emph{both the hand and the object}, using
conditioning~\cite{unidiffuser,ugg,semgrasp,text2grasp,diffh2o,handiffuser,ghop}).
While these methods have useful applications, they cannot support newly emerging applications.
For instance, in virtual reality, we may want to automatically synthesize 3D held objects based on user's hand pose/shape~\cite{virtual_object1, virtual_object2}.
In generative image editing, we may want to edit an image of a hand~\cite{handiffuser} to either add or replace a (held) object.

To support these emerging applications, we consider the inverse problem: \emph{given a 3D model or 2D image of a hand (either free or holding another object), could we synthesize a \textbf{held object} that fits the input hand pose?}
This previously underexplored~\cite{objectpopup} problem setting is highly challenging.
First, the space of 3D object variation is significantly larger than that of hands, making it hard to synthesize plausible and diverse objects.
Second, the contact and occlusions, and the requirement to keep hand pose unchanged, make it challenging to synthesize accurate held objects.
Finally, despite incredible progress in image editing~\cite{stable_diffusion, ddim, cfg,imagic, instructpix2pix, editable_image_elements, sdedit, paintbyexample}, generating realistic images of hands and grasps has been found to be particularly hard~\cite{WhyAIart15:online, Criminals:online} (see ~\Cref{fig:SD_inpainting}).

To address these challenges, we present \textbf{\shortname}, a method that synthesizes 3D or 2D \textbf{held objects} conditioned on a 3D hand model or a 2D image of a hand.
Given the 3D pose and/or shape of a hand (e.g.,~MANO~\cite{mano}), our \textbf{\shortnamethree} module is capable of selecting a plausible held object from the large and diverse Objaverse dataset~\cite{objaverse}, and posing the object so that the hand grasps it without articulation changes.
To enable this, \shortnamethree consists of an object selection network that is trained on a dataset of grasps~\cite{dexycb} to generate \textbf{object codes} -- compact representations of the geometry of plausible held objects.
These codes are then matched to objects in the Objaverse dataset to select a suitable held object.
To ensure that the selected held objects can be grasped without change in hand pose, we introduce an object fitting procedure that optimizes for contacts, interpenetration, and physical laws.

If only a 2D image of a hand, either free or holding an object, is available, we can still handle it with \textbf{\shortnametwo}.
\shortnametwo builds on top of \shortnamethree, but combines it with the editing capabilities of diffusion models~\cite{stable_diffusion}.
Given a hand image, we first extract 3D pose and shape~\cite{hifihr, nimble} and use \shortnamethree to generate a plausible 3D held object.
The 3D hand and generated held object are used as 3D guidance for a diffusion-based image editing process to add or replace existing held objects in the input image (see ~\Cref{fig:teaser}).
We compare 3D grasp quality, object plausibility, and 2D editing plausibility of our method with related work.
We show that the ability to select the plausible object greatly improves the 3D grasping quality , and leads to faster convergence in the object fitting.
The capability of using an occlusion-aware 3D grasp as guidance demonstrates superior 2D editing plausibillity over not using such guidance (e.g., guided by text).
To summarize our contributions:
\begin{packed_itemize}
%
    \item We study a previously underexplored grasping problem: given a 3D model or 2D image of a hand, we want to synthesize a \textbf{held object} that fits the input hand pose.
    \item We introduce \textbf{\shortnamethree}, a module that selects a plausible 3D held object from a large and diverse dataset using object codes and ensures physical grasp plausibility.
    %
    \item We introduce \textbf{\shortnametwo}, a module that can add or replace to an image of a hand by combining the editing capabilities of diffusion models with \shortnamethree.
    %
\end{packed_itemize}

\section{Related Works}
\label{sec:related_works}
\vspace{-0.2cm}
\paragraph{Grasp Synthesis.}
Grasp synthesis refers to the task of generating either pose of a human or robot hand such that an object is grasped.
In computer vision and graphics, the focus of grasp generation has been on plausible grasps that make accurate contact, without interpenetration, and physically correct~\cite{dgrasp,graspd, grasping_fields, grasptta, goal, saga, toch, cams}. 
In robotics, however, the focus has been more on \emph{stable grasps} that result in objects being successfully grasped in simulation or the real world~\cite{unidexgrasp++, gendexgrasp, se3_diffusionfields, dexgraspnet, dexdiffuser, crosshand}.
Many methods take 3D object representations as input for grasp synthesis~\cite{grasptta, grasping_fields, toch, goal, dgrasp, graspd, contactgen, contactedit, cams, hoidiffusion, graspxl, realdex}.
Heuristics~\cite{graspit} and physical simulation~\cite{grasptta,dgrasp,graspd} are common ways of optimizing for stable grasps at inference time.
ContactGen~\cite{contactgen} predicted plausible contacts on objects, while HOIDiffusion~\cite{hoidiffusion} generated new images from synthesized grasps.
Other approaches take an image as input and generate hands on the 2D image~\cite{affordancediffusion,handiffuser,ghop}.
The proliferation of large language models has also inspired methods to generate hands grasping objects from text prompts~\cite{semgrasp, text2grasp, diffh2o}.

\begin{wrapfigure}[13]{r}{0.65\textwidth}
    \centering
    \vspace{-0.2in}
    \includegraphics[width=1.0\linewidth]{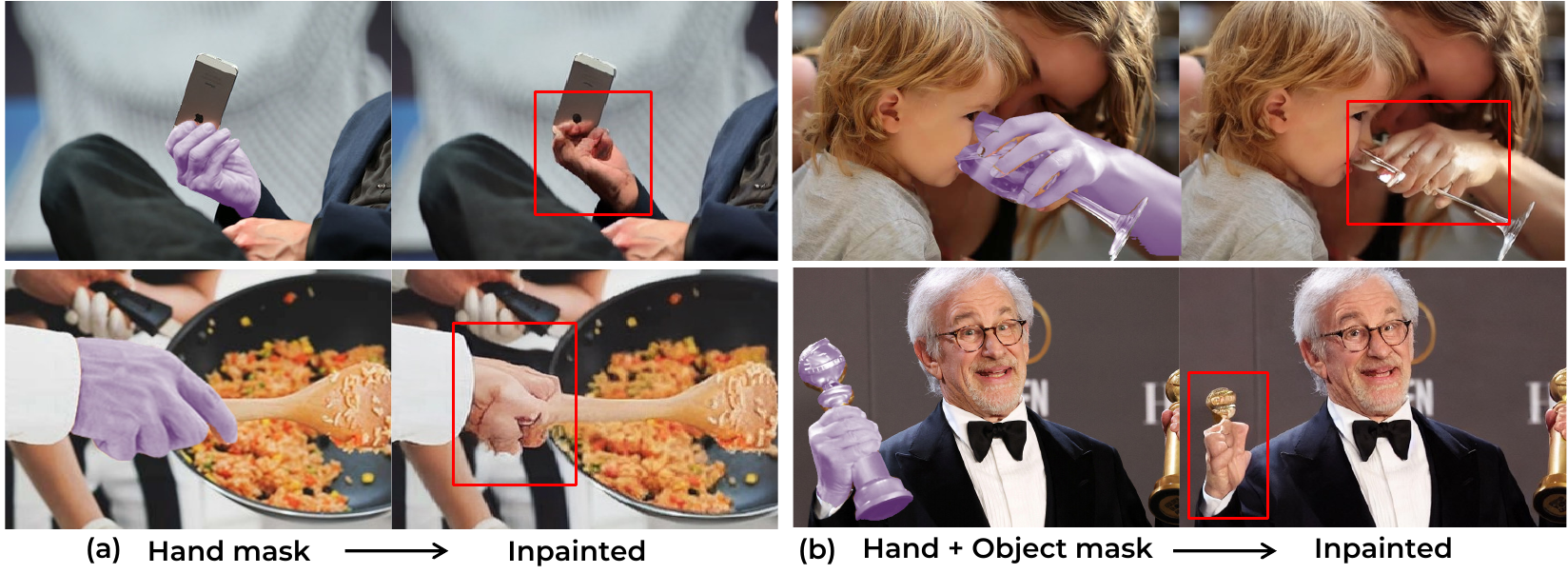}
    \caption{
    Stable Diffusion~\cite{stable_diffusion} struggles to edit images of hands holding objects.
    For an identity inpainting task (purple region), it is unable to faithfully reconstruct the hand or held object.
    }
    \label{fig:SD_inpainting}
    \vspace{-.5cm}
\end{wrapfigure}

Numerous datasets~\cite{obman, affordpose, dexgraspnet, ddgdata, diva}, simulation environments~\cite{graspit, issac_gym}, and hand/objects datasets~\cite{shapenet, partnet, mano, manus} have enabled research in grasp synthesis.
This has enabled the study of real-world grasps with one hand~\cite{dexycb}, two hands~\cite{grab, behave, ho3d, arctic, contactpose}, or robot hands~\cite{realdex} approaching and finally grasping objects.
In this paper, we take advantage of the DexYCB~\cite{dexycb} to learn object codes that enable us to select suitable objects from a large dataset~\cite{objaverse}.
The grasping studies also get much attention from robot grippers and the full-body motion generation. 
For robotics applications, ~\cite{unidexgrasp++, gendexgrasp, se3_diffusionfields, dexdiffuser} considered physical laws and experimented on simulators.
One such physics-aware approach~\cite{forceclosure} allows the grasping simulation to be differentiable, by formulating the force closure loss.
We integrate this method in our object fitting procedure.

Some previous work integrate the entire body in the grasping process~\cite{saga, grab, goal, objectpopup, grip}.
Object Pop-up~\cite{objectpopup} is a closely related work that generates plausible objects given a 3D point clouds of entire human body.
However, our work differs in that we can generate more accurate contacts of objects from only 3D hand models.
Furthermore, we also support direct editing of 2D hand images. 

\paragraph{Image Editing.}
Editing images is a long-standing problem. 
For the object pasting and blending task, classical works included the famous alpha blending~\cite{alpha_blending} and poisson blending~\cite{poisson_blending}. 
Recently, the quality of image generation diffusion models~\cite{ddpm, ddim, cfg} has enabled realistic editing capabilities.
Such models include object resizing~\cite{editable_image_elements, geodiffuser}, moving~\cite{editable_image_elements, geodiffuser, diffusionhandles, dragondiffusion}, removal~\cite{editable_image_elements, geodiffuser}, variation~\cite{editable_image_elements}, and editing by text~\cite{prompt2prompt, imagic, instructpix2pix}. 

In this paper, we use \shortnamethree to generate plausible 3D objects and 3D hand pose/shape to guide editing of the original image.
This leads us to focus on the object pasting which accepts reference image and source image. 
Paint-by-Example~\cite{paintbyexample} and DiffEditor~\cite{diffeditor} support such object pasting. 
Paint-by-Example~\cite{paintbyexample} allows changes in the reference image so that it can semantically harmonizes with the source image. 
On the other hand, DiffEditor~\cite{diffeditor} leverages the DDIM inversion~\cite{ddim} and score-based guidance~\cite{score_based1, score_based2} with SDE~\cite{ddpm} to balance between flexibility and preservation.
Some methods focus on editing images, using text prompts only. 
Imagic~\cite{imagic} edits real images, based on the text description of the edited result.
InstructPix2Pix~\cite{instructpix2pix} edits real images with only the prompt about the edited part. 
While these methods can perform image editing, using only the text prompts may lead to difficulties in a particular problem -- editing held objects. 
In experiments, we show that the \shortnametwo (combining the \shortnamethree into the existing object pasting pipeline) leads to better results than these methods. 
Notably, many of the aforementioned image editing techniques are general-purpose models, and thus are not naturally guided by any additional priors, e.g., 3D shapes and the hand template models~\cite{mano, nimble}. 
On the contrary, our approach is specifically designed for the hand grasping setting.
%

\section{Method}
\label{sec:method}
\vspace{-0.3cm}
Our proposed method, \shortname, is designed to support two mains tasks.
\shortname3D takes 3D hand pose/shape as input and produces a plausible 3D held object.
\shortname2D takes a hand image as input and edits the image to add or replace a held object using 3D guidance from \shortname3D.
In the following sections, we described each module in detail.

\begin{figure}[ht!]
    \centering
    \includegraphics[width=0.9\textwidth]{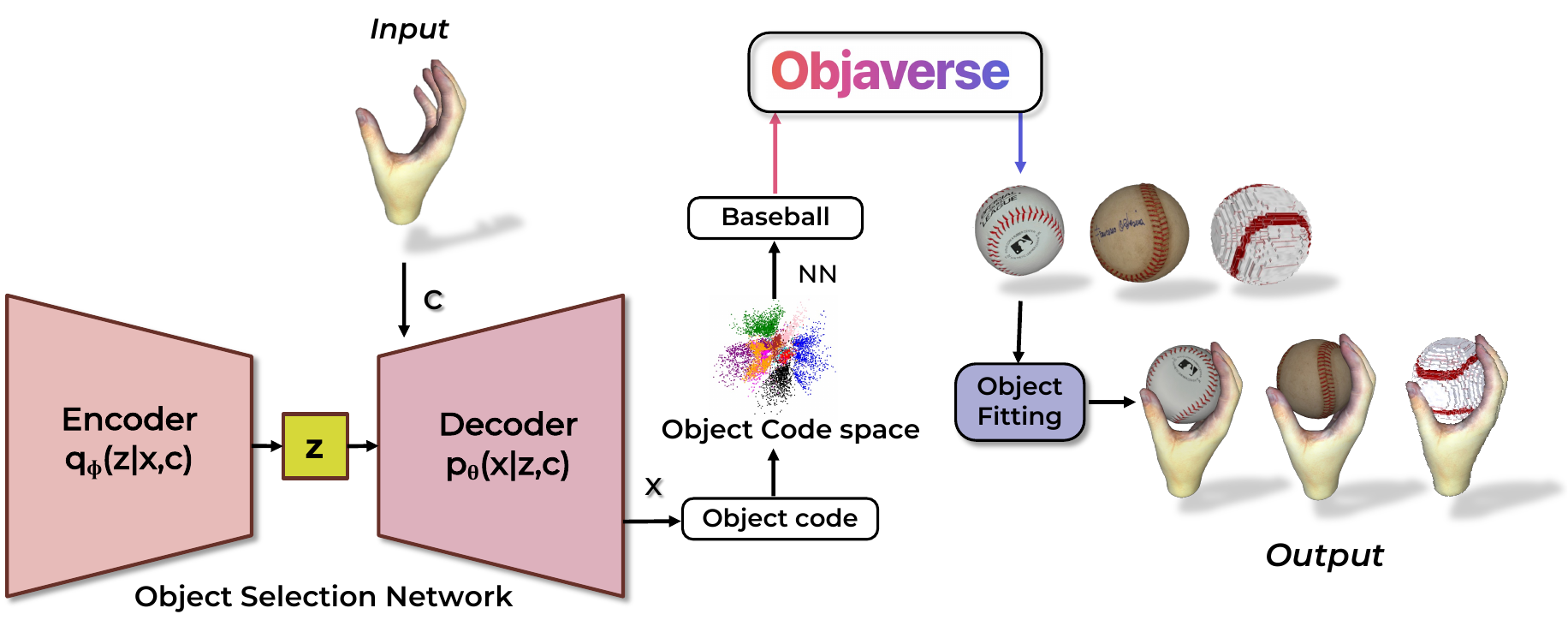}
    \vspace{-0.2in}
    \caption{\shortnamethree can synthesize a 3D held object given a 3D hand model as input (top).
    We encode the 3D hand model to estimate \textbf{object codes} that act as a compact representation of plausible held objects.
    These object codes can be used to retrieve diver objects from a much larger dataset like Objaverse~\cite{objaverse}.
    This is followed by an object fitting step to position and orient objects to form the grasp without changing the initial hand pose.
    }
    \label{fig:main_arch}
\end{figure}
\subsection{GenHeld3D}
\label{subsec:genheld3D}
\vspace{-0.2cm}
When given a 3D hand model (pose and shape) as input, the goal of \shortname3D is to select an appropriate object from a large dataset of objects and pose it in a plausible grasp without changing the pose of the hand.
To achieve this, we split the task into two parts: object selection and object fitting.

\paragraph{Object Selection.}
\label{subsubsec:object_selection}
Synthesizing plausible and diverse objects corresponding to an input hand requires learning from real-world grasps of objects.
However, existing grasp datasets~\cite{ho3d,behave,grab,dexycb,dexgraspnet,arctic,affordpose,unidexgrasp++,realdex} are limited in the diversity of held objects.
We seek to select plausible and diverse held objects from large object datasets like Objaverse~\cite{objaverse}.
To achieve this, our main insight is to learn \textbf{object codes} -- low-dimensional encodings of the object shapes from existing grasp datasets which can guide the selection of objects from a large object database.
%

We define an object code as $\bold{x} = [\frac{l_3}{b}, \frac{l_2}{l_3}, \frac{l_1}{l_2}]$, where $l_i$ are the three lengths of tight 3D object bounding box. 
$l_i$ are obtained by extracting principal components~\cite{pca} of object vertex positions, and sorted to ensure the rotational invariance. 
We also define principal bone length $b$ as the the distance between root and the first joint of thumb, defined in MANO~\cite{mano}. 
We aim to learn the distribution of $\bold{x}$ from the 20 object categories~\cite{ycb} in a dataset of real hand grasps~\cite{dexycb}.
We can then use the learned object code distribution to find nearest matching category in a larger dataset~\cite{objaverse} and sample diverse instances from that category.
Since datasets like Objaverse can have object instances with arbitrary scales, we can rescaling
the bounding box lengths of the fetched object 
as $l_1=x_1 b, l_2=x_1x_2b, l_3=x_1x_2x_3b$.
\Cref{fig:shape_code} visually describes the object selection procedure.

\paragraph{Predicting Object Codes.}
%
Our main insight is that object codes are associated with hand pose/shape.
Thus, it should be possible to learn the association between 3D hand models and object code from existing grasp datasets.
To do this, we bulid an object selection network (see \Cref{fig:suppl_arch_detail}) that predicts object codes from an 3D hand model.
During inference, this network takes as input 3D hand pose and shape (vertices) and outputs object code and 3D contacts $\mathbf{h_c}$.
We find that the contact plays an important role, because it is in the intersection between hand and object is made by contact.
During training, a point cloud of the object shape, ground truth object codes and contact are provided as supervision.
Please see the supplementary document for more details.

%
In summary, we try to bridge the object shape code generation task with hand pose, object, and contact.
Shape code, object point clouds, and contact is supervised during training. 
For the ground truths, we acquire the paired data from an external dataset~\cite{dexycb}.
Also, we acquire the $h_c^*$ by thresholding the distance between the vertices of hand and the object. 
The network is trained by the loss 
$\mathcal{L}_\text{select} = \mathcal{L}_\text{rec} + \lambda_\text{KL}\cdot\mathcal{L}_\text{KL}$, and $\mathcal{L}_\text{rec}$ is given by,
\begin{equation}
    \mathcal{L}_\text{rec} = ||\bold{x} - \bold{x}^*||_2 + \lambda_o\cdot||{O} - O^*||_2 + \lambda_\text{con}\cdot \text{BCE}(\bold{h}_\bold{c}, \bold{h}^*_\bold{c}),
\end{equation}
where $^*$ denotes the ground-truths.
$\lambda$ controls the balances between losses. 
The KL divergence term is given by,
\begin{equation}
    \mathcal{L}_\text{KL} = \text{KL}[\mathcal{N}(\mu_\bold{z}, \Sigma_\bold{z}^2)||\mathcal{N}(0, I)].
\end{equation}
We apply an annealing strategy~\cite{posterior_collapse1, posterior_collapse2} to $\mathcal{L}_\text{KL}$ in order to avoid posterior collapse.
During inference, we sample a latent vector $\bold{z}$ to feed the decoder $p_\theta$.

\begin{figure}
    \centering
    \includegraphics[width=1.0\textwidth]{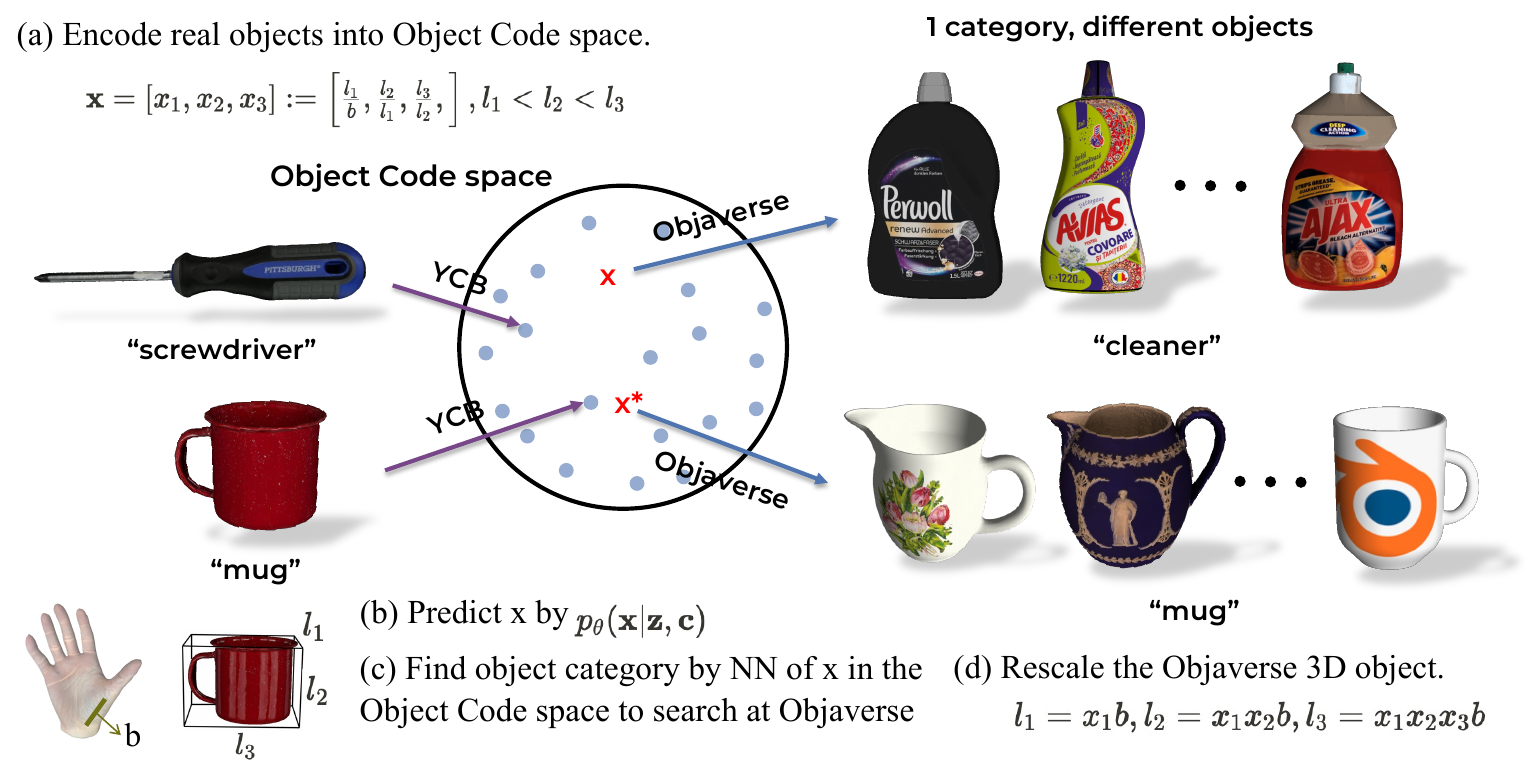}
    \caption{
    We use \textbf{object codes} -- compact representations of object shapes learned from a real dataset of grasps.
    These codes can be used to find suitable diverse objects in a larger object dataset like Objaverse.
    Our method can also handle scale variations by normalizing using principal bone length $b$.
    }
    \label{fig:shape_code}
    \vspace{-.5cm}
\end{figure}

\paragraph{Object Fitting.}
\label{subsubsec:object_fitting}
When the fine-grained 3D object is fetched from Objaverse~\cite{objaverse}, our method tries to fit the object to the given hand.
Note that he hand should be preserved, so only the object pose should be optimized. 
To achieve this, we incorporate multiple losses that control contact, penetration, and physical laws.
We optimize the object transformation using the following loss function
%
\begin{equation}
    \mathcal{L}_\text{opt} = \lambda_A \mathcal{L}_A + \lambda_R \mathcal{L}_R + \lambda_\text{sim} \mathcal{L}_\text{sim},
\end{equation}
where $\lambda$ controls the strengths between terms.
%
The attraction loss $\mathcal{L}_A$, and repulsion loss $\mathcal{L}_R$ are given as,
\begin{equation}
    \mathcal{L}_A = \sum_{i=1}^6 \Phi_\alpha (d(C_i \cap \textsc{Pene}_\bold{h}(V_\text{obj})^c, V_\text{obj})), \quad \mathcal{L}_R = \sum_{v\in \textsc{Pene}_\bold{h}(V_\text{obj})} \Phi_\alpha (d(v, V_\text{obj})).
    \label{eq:fitting_loss}
\end{equation}
We follow the specific formulation produced by~\cite{obman}, but we only allow object translation, rotation, and scaling as variables. 
The object scaling is wrapped by a Sigmoid function to avoid scale explosion, as $s(x) = \frac{2k}{1+\text{exp}(-x+1)}+1-k$, where we allow the max scale to be $1+k$ and min scale to be $1-k$. 
We use the common penalization function $\Phi_\alpha (x) = \alpha \text{tan}(\frac{x}{\alpha})$ with the characteristic distance action $\alpha$. 
The $C = \left\{C_i\right\}, i \in [1, 6]$ is given by the statistical set of contact vertices from~\cite{obman}. 
The minimum distance between a vertex point and a set of vertices is defined as $d(v, V) = \text{inf}_{w\in V} ||v - w||_2$, and the minimum distance between the two sets of vertices is given by $d(\Tilde{V}, V) = \text{inf}_{v\in\Tilde{V}}d(v, V)$. 
Plus, we compute the penetration ($\textsc{Pene}$) with the classic Ray-Triangle intersection~\cite{ray_tri_intersec} for better efficiency. 
%
Inspired by the Differentiable Force Closure~\cite{forceclosure, forceclosure_original}, we also enforce the
physical plausibility using a differentiable simulation loss
$\mathcal{L}_\text{sim}$.
This makes sure the forces exerted on the object is closed, so that the object will not fall down.
Please see more details in the supplementary \Cref{sec:suppl_simloss}.

\subsection{GenHeld2D}
\label{subsec:genheld2D}
While \shortnamethree can generate held objects given 3D hand model, we often may need to add or replace held objects in a hand image.
We show how \shortnamethree can be combined with the editing capabilities of diffusion models to build \shortnametwo that operates directly on images (see \Cref{fig:2D_arch}).
\begin{wrapfigure}[21]{r}{0.6\textwidth}
    \centering
    \includegraphics[width=1.0\linewidth]{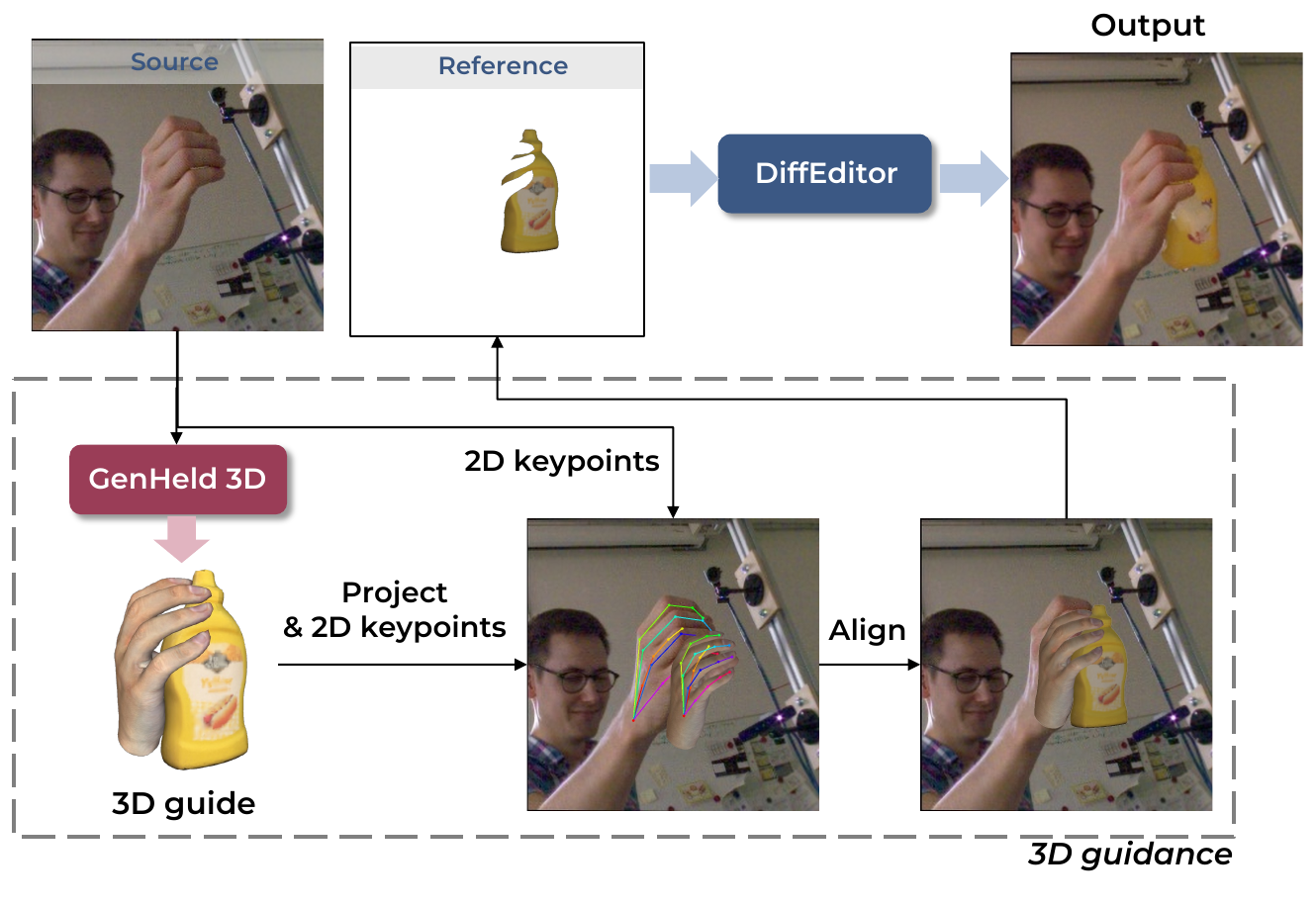}
    \vspace{-.5cm}
    \caption{\shortnametwo enables us to add or replace held objects to 2D hand images.
    We do this by first lifting hand images to 3D hand and object using \shortnamethree.
    This is then following by 2D keypoint projection and alignment to create a 3D guidance image that is used to edit the image.
    }
    \label{fig:2D_arch}
    \vspace{-.3cm}
\end{wrapfigure}

\paragraph{3D Guiding.}
\label{subsubsec:image_editing}
The first step in operating on images is to obtain a 3D hand model corresponding to the input image.
We then use this to generate a 3D held object using \shortnamethree. 
As shown in \Cref{fig:2D_arch}, we first project the result of \shortnamethree onto the source image.
Since the hand and object were defined in 3D, we can benefit from the pixel-wise information about the z-buffer. 
That means we can render the hand and object, either together, independently, or separately in a occlusion-aware way. 
This allows us to rectify the potential misalignment by leveraging the hand 2D keypoints from the independently projected hand.
Also, such information on z-buffer enables to render only the un-occluded object parts onto the source image.
By doing this, we can obtain the reference image, as depicted in \Cref{fig:2D_arch}. 

\paragraph{Object Pasting.}
%
Since \shortnamethree is able to provide a 3D guidance in the form of occlusion-aware grasped object and the source image.
We then use, DiffEditor, a state-of-the-art object pasting model~\cite{diffeditor} without needing any training or fine-tuning.
First, we perform DDIM Inversion~\cite{ddim} on both the source and reference image.
This process produces keys and queries of those real images, which are fed into the cross attention module.
Additionally, we guide the reverse process by the score-based function $\nabla_{x_t} \text{log} q(x_t | y) \propto \nabla_{x_t} \text{log} q(x_t) + \nabla_{x_t} \text{log} q(y | x_t)$, introduced in \cite{score_based1, score_based2}.
y denotes the condition and $x_t$ defines the image at timestep $t$.
The last term implies the conditional gradient, while the first term implies the unconditional denoiser. 
Complex finger articulations make the preservation of the fingers a challenging problem. 
As a result, we need an extra guidance from the mask.
Thus, we additionally incorporate the regional mask-aware gradient guidance, proposed by~\cite{dragondiffusion}, on the conditional gradient.
Such guidance is given by,
\begin{equation}
    \nabla_{z_t} \text{log} \; q(y|x_t) = \bold{m}_\text{edit} \cdot \nabla_{x_t} \mathcal{E}_\text{edit} + (1-\bold{m}_\text{edit})\cdot \nabla_{x_t}\mathcal{E}_\text{content},\nonumber
    \label{eq:score_based}
\end{equation}
where the mask $\bold{m}_\text{edit}$ lies in the edited region, and $\mathcal{E}$ is the energy functions.
We provide more details in the supplementary \Cref{sec:suppl_2D}.
Moreover, we observe that it is easy to get a result that is not edited sufficiently, if we solely rely on the ODE~\cite{ddim}, e.g., ~\cite{dragondiffusion}. 
This phenomenon on grasping editing problem encouraged us to adopt the interval-timestep combination, inspired by ~\cite{diffeditor}, of SDE~\cite{ddpm} and ODE~\cite{ddim} to involve more flexibility without harming the overall context. 
Then, the sampling is as follows,
\begin{equation}
    x_{t-1} = 
    \begin{cases}
        \bold{m}_\text{edit} \cdot \mathcal{F} (x_t; \eta_1) + (1-\bold{m}_\text{edit})\cdot \mathcal{F} (x_t; \eta_2) &  , T-t < n \; \text{and} \; t \% 2 = 0 \\
        \mathcal{F} (x_t; 0) & , \text{otherwise},\nonumber
    \end{cases}
\label{eq:diffeditor_interval}
\end{equation}
where $\eta_1 > \eta_2$.
$\mathcal{F}$ means the non-Markovian process in DDIM~\cite{ddim}, described in ~\Cref{sec:suppl_2D}.
Following \Cref{eq:DDIM_step} (in supplementary), when $\eta \neq 0$, the process becomes stochastic (SDE) with the random noise involved (otherwise, ODE).

\paragraph{Object Replacement.}
%
\shortnametwo can also be used to replace an object in people's hands rather than adding a new one.
In this setting we perform hand segmentation~\cite{egohos,sam}, remove the object and inpaint the background~\cite{stable_diffusion}.
Although, SD inpainter is prone to fail when human hands are involved (Fig.~\ref{fig:SD_inpainting}), we find that inpainting only the object does not degrade the results.
After inpainting, we perform the same pipeline as the Object Pasting process.

\paragraph{Non-Grasping Hand Pose Rejection.}
Throughout the pipeline, we assume that the input hand image shows a grasping pose -- either with or without object. 
Previously, UGG~\cite{ugg} and DexGraspNet~\cite{dexgraspnet} suggested a method to evaluate the generated hand pose by deciding whether the hand is grasping or not.
Inspired by this, to filter out the potential non-grasping hand image input, we propose a method using the ConvexHull as described in the supplementary (\Cref{sec:suppl_handrejection}). 
\vspace{-0.2cm}
\begin{figure}[ht]
    \centering
    \includegraphics[width=0.9\textwidth]{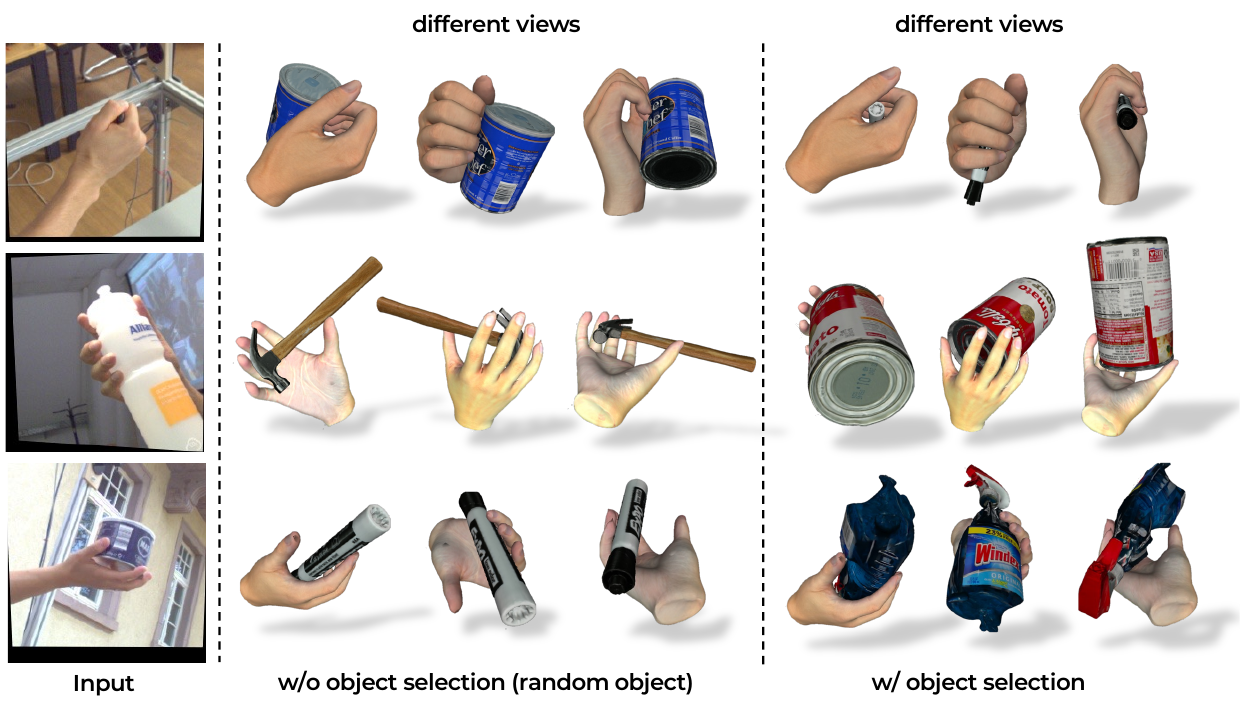}
    \caption{Ablation study of the Object Selection Network with YCB~\cite{ycb} objects.
    We observe that object selection is an important step to ensure the plausibility of generated held objects.
    }
    \vspace{-0.2cm}
    \label{fig:ablation}
\end{figure}

\section{Experiments}
\label{sec:experiments}
\vspace{-0.2cm}
\begin{wraptable}[10]{r}{0.5\textwidth}
    \centering \small
    \begin{tabular}{lccc}
         \toprule
                                  & SD$\downarrow$   & Pene depth$\downarrow$ & Pene vox.$\downarrow$ \\
                                  & [cm]             & [cm]                   & [cm$^3$]              \\
         \midrule
         GraspTTA~\cite{grasptta} & 3.21             & 1.05                   & 4.58                  \\
         Cpf~\cite{cpf}           & 2.77             & 1.14                   & 3.46                  \\
         DGrasp~\cite{dgrasp}     & 1.9              & -                      & -                     \\
         Ours                     & \textbf{1.23}             & \textbf{1.11}                   & \textbf{3.02}                  \\
         \bottomrule
    \end{tabular}
    \caption{Grasping quality comparison.
    \textbf{\shortnamethree} outperforms existing work.}
    \label{tab:grasping_quality}
\end{wraptable}

In this section, we provide qualitative and quantitative results, and justify key design decisions.
For more details, please see the Appendix.

\begin{wraptable}[6]{r}{0.48\textwidth}
    \centering \small
    \begin{tabular}{lcc}
         \toprule
                                & w/o selection    & w/ selection     \\
         \midrule
         \# of iters$\downarrow$& 3235.9 (N/A 53)  & 1837.3 (N/A 21)  \\
         \bottomrule
    \end{tabular}
    \caption{Object Selection plausibility.
    We see a singnificant reduction in object fitting time when the selection network is used.
    }
    \label{tab:selection_plausibility}
\end{wraptable}

\paragraph{Datasets and Setting.}
We trained our Object Selection Network from scratch with the DexYCB~\cite{dexycb} dataset which consists of 582K frames with 10 real human subjects and 20 real objects with accurately known sizes.
For 2D input images, we used the test images (4K) from the FreiHAND~\cite{freihand} dataset for overall experiments and HO3D~\cite{ho3d} in \Cref{tab:grasping_quality} for fair comparison.  
For building the object code space, we use an existing graspable object dataset~\cite{ycb}.
For the object fitting, we use the Adam~\cite{adam} optimizer and set the max number of iterations as 4K. 

\subsection{\shortnamethree Results}
\label{subsec:3D_results}
%
In this section, we evaluate the grasp quality (\Cref{tab:grasping_quality}) and object plausibility (\Cref{tab:selection_plausibility}) of \shortnamethree. 
For the Object Selection network, we perform an ablation study in ~\Cref{tab:grasping_quality_ab}, by using the grasping quality metrics. 
We also show visual results in \Cref{fig:ablation}.

\paragraph{Evaluation Metrics.}
For the grasping quality, we employ the common metrics with the exactly same hyper-parameters with GraspTTA~\cite{grasptta}, Cpf~\cite{cpf}, and DGrasp~\cite{dgrasp}. 
We used the PyBullet~\cite{pybullet} simulator for the Simulation Distance (SD).
We also compute the maximum penetrating depth and the penetrating volume of the discretized voxels that are penetrating. 
The average hand size was 229.947~cm$^3$.
For the object selection plausibility, we measure how the selected object matches the hand pose, by counting the number of object fitting steps until the loss reaches half the initial loss. 
At (N/A), we put the number of instances that failed to reach the halving loss until the maximum number of iterations.

\begin{table}[ht]
    \centering \small
    \begin{tabular}{lcccc}
        \toprule
                        & SD[cm]$\downarrow$ & Success ratio$\uparrow$ & Pene depth[cm]$\downarrow$ & Pene vox.[cm$^3$]$\downarrow$ \\
        \midrule
        w/o selection   & 1.298$\pm$1.698    & 0.604$\pm$0.489         & 0.316$\pm$0.436            & 4.541$\pm$6.276              \\
        w/o shape code &1.341$\pm$1.884&0.690$\pm$0.433&0.312$\pm$0.530& 4.003$\pm$5.239\\
        w/o $O$ &1.090$\pm$1.342&0.645$\pm$0.613&0.238$\pm$0.337&4.424$\pm$5.011\\
        w/o $\bold{h}_c$ &1.203$\pm$2.015&0.432$\pm$0.587&0.180$\pm$0.509& 2.908$\pm$5.219\\
        Ours (full)  & \textbf{0.791}$\pm$0.829    & \textbf{0.791}$\pm$0.829         & \textbf{0.178}$\pm$0.490            & \textbf{2.687}$\pm$5.456              \\
        \bottomrule
    \end{tabular}
    \caption{Ablation study of the Object Selection network.
    Object selection, shape code, object point cloud and contacts are all essential for best performance.
    }
    \label{tab:grasping_quality_ab}
\end{table}

\vspace{-0.3in}
\paragraph{Analysis.} 
\Cref{tab:grasping_quality} shows the ability of our object selection and object fitting results in high quality in grasping. 
The ability of the Object Selection network is further demonstrated in Tab.~\ref{tab:selection_plausibility}. 
When the Object Selection network is applied, our object fitting shows a significantly faster convergence. 
This signifies that the predicted objects were plausible enough to ease the grasping process.
The necessity of the Object Selection is also illustrated in Fig.~\ref{fig:ablation}.
The ablation study of Object Selection network and other components is shown in Tab.~\ref{tab:grasping_quality_ab}. 
The w/o shape code is when we make the Object Selection network directly predict a object category and re-scale the Objaverse~\cite{objaverse} object by its radius. 
Only when all the components claimed in our architecture are given, we get our best results.

\begin{wraptable}[9]{l}{0.65\textwidth}
    \centering \small
    \begin{tabular}{lccc}
         \toprule
                                                       & Imagic~\cite{imagic} & Pix2Pix~\cite{instructpix2pix}& Ours   \\
         \midrule
         (a) FID $\downarrow$                          & 155.83               & 162.2                         & 135.45 \\
         (b) CLIP score (prompt) $\uparrow$            & 0.24                 & 0.26                          & 0.25   \\
         (c) CLIP score \textit{"realistic"} $\uparrow$& 0.20                 & 0.19                          & 0.28   \\
         (d) masked PSNR [dB] $\uparrow$               & 29.055               & 31.875                        & 34.002 \\
         (e) Time $\downarrow$                         & 20m 25s              & 15s                           & 1m 48s \\
         \bottomrule
    \end{tabular}
    \vspace{-0.1in}
    \caption{(a-c) Image fidelity , and (d) hand preservation of \textbf{\shortnametwo}}
    \label{tab:2D_image}
\end{wraptable}

\subsection{\shortnametwo Results}
\label{subsec:2D_results}

In this section, we evaluate the \shortnametwo. 
In tab,~\ref{tab:2D_image}, we compare the image quality and hand preservation in the result images.
Fig.~\ref{fig:qualitative} shows the held object addition results. 
Fig.~\ref{fig:object_replacement} demonstrates the held object replacement results.

\begin{figure}[t!]
    \centering \footnotesize
    \includegraphics[width=\textwidth]{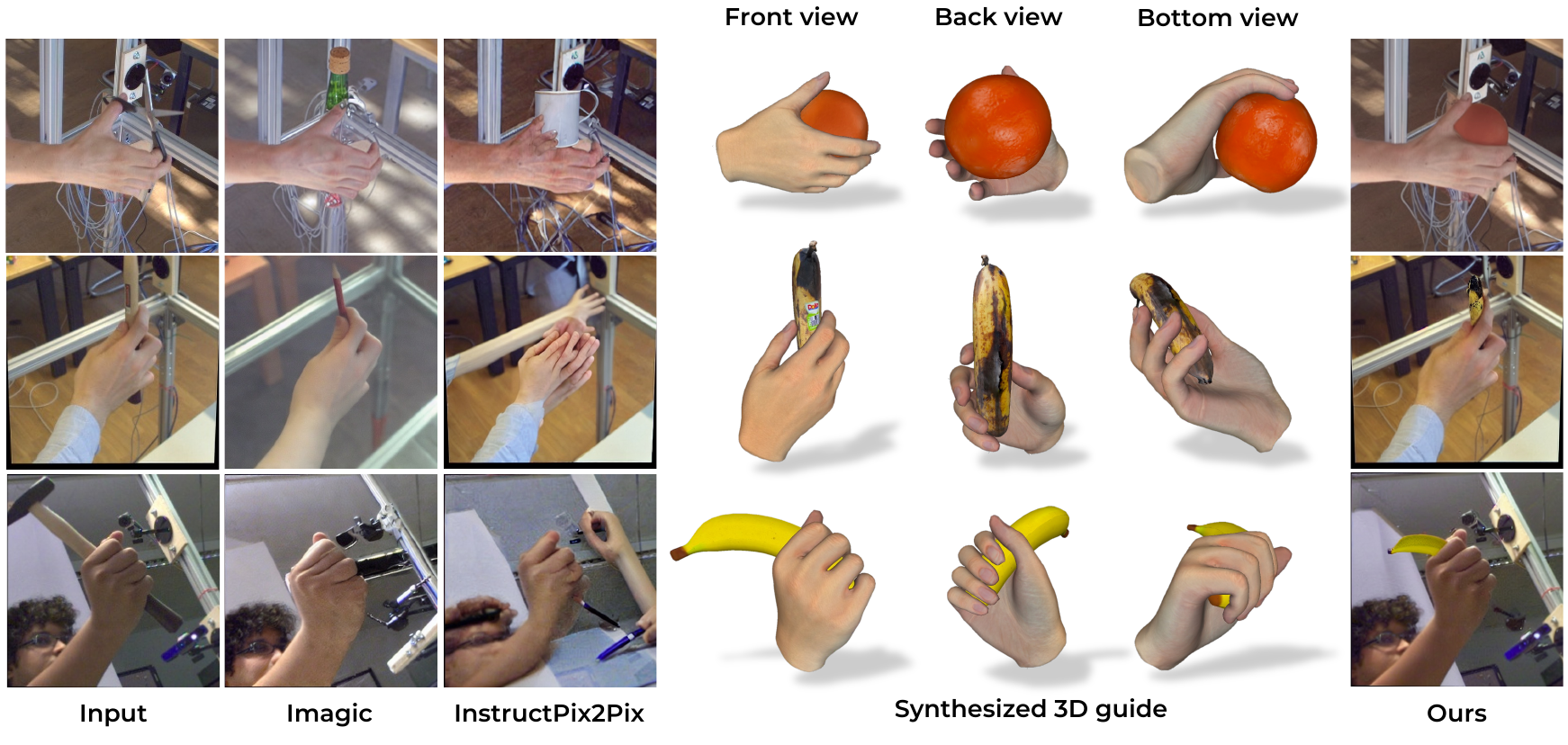}
    \vspace{-0.1in}
    \caption{The object replacement application. For Imagic~\cite{imagic} and InstructPix2Pix~\cite{instructpix2pix} which can only rely on text prompts, we gave \textit{"Replace the object in the hand with another object"}.}
    \vspace{-0.25in}\label{fig:object_replacement}
\end{figure}

\begin{figure}[b!]
    \centering
    \includegraphics[width=\textwidth]{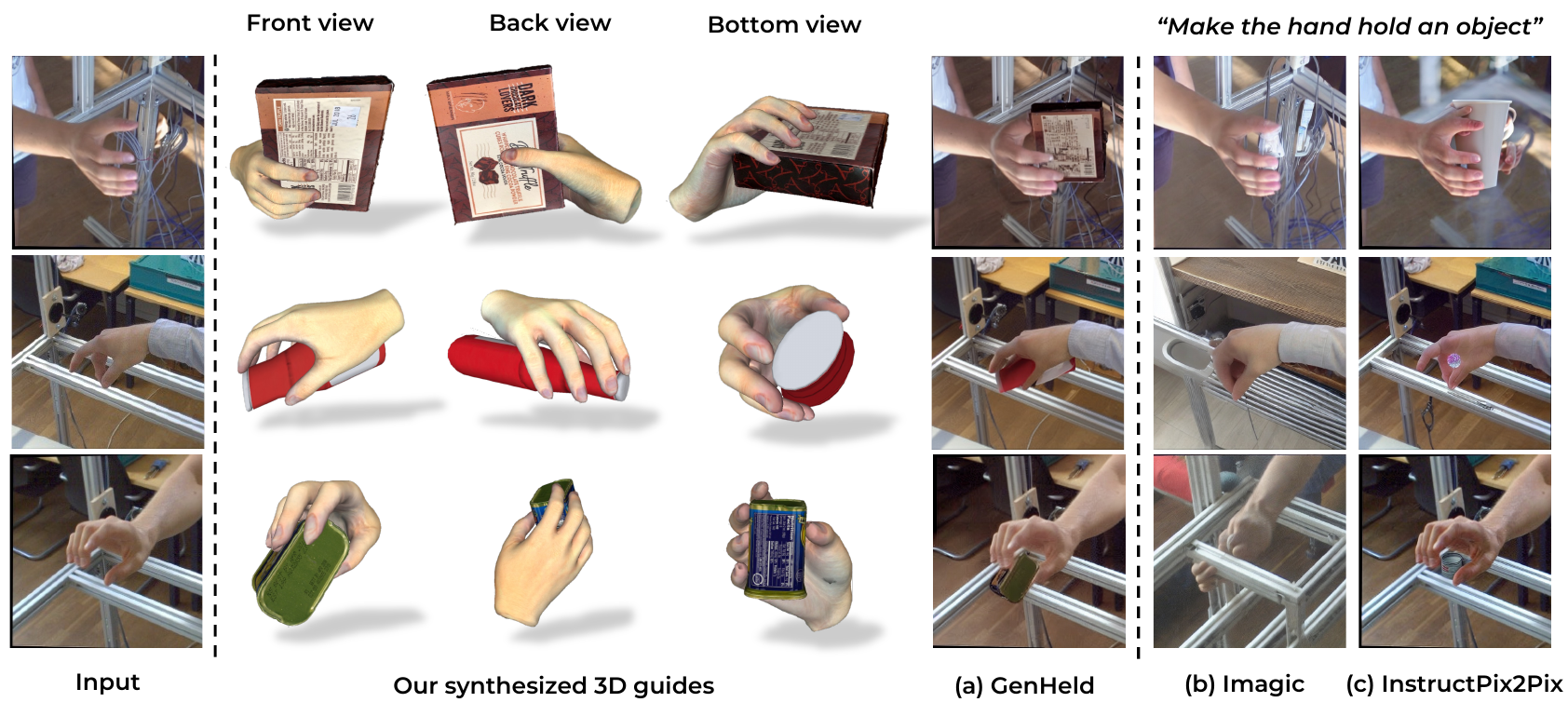}
    \caption{Qualitative comparison with the Imagic~\cite{imagic} and the InstructPix2Pix~\cite{instructpix2pix} which can only rely on text prompts.}
    \vspace{-0.2cm}
    \label{fig:qualitative}
    \vspace{-0.3cm}
\end{figure}

\textbf{Evaluation Metrics.}
First, we compare the distance between the features of the FreiHAND testset images and those of our synthesized images by FID~\cite{fid}.
Roughly half of the FreiHAND testset images are empty hands, and the other half are hands with object. 
We used the second half only, when we compute the FID score. 
For the CLIP~\cite{clip} score, we evaluted how the synthesized images align with the prompt "Hand holding an object".
We assume that the relatively poor understanding of general-purpose models resulted in the minimal differences between methods in Tab.~\ref{tab:2D_image}.
Therefore, we introduce another measure, which is CLIP score on the text "realistic", to additionally evaluate the realisticity of the generated images.
Lastly, we propose the masked PSNR. 
We use the ground-truth hand mask provided by FreiHAND~\cite{freihand} and compute the PSNR between the input and output images in the masked regions.
By doing this, we try to measure the preservation of the hand in the edited images.

\textbf{Analysis.}
We show the generated image quality in Tab.~\ref{tab:2D_image}. 
Overall, our method shows a significant elevation of (a-c) the edited image fidelity and (d) the hand preservation.
Fig.~\ref{fig:qualitative} compares the qualitative results between ours, Imagic~\cite{imagic}, and InstructPix2Pix~\cite{instructpix2pix}. 
~\cite{imagic} and ~\cite{instructpix2pix} both shows artifacts in fingers and often fails to make the hand holding a new object. 
On the other hand, with the 3D guides, we can achieve the results with higher fidelity. 
Finally, we showcase the Object replacement application in Fig.~\ref{fig:object_replacement}.
Similar with the object generation, our paradigm of automatic category prediction and the grasping optimization in 3D significantly alleviate the artifacts and unrealistic hand-object shown in the existing methods~\cite{imagic, instructpix2pix}.

\vspace{-.5cm}
\section{Conclusion}
\label{sec:conclusion}
\vspace{-.2cm}
We introduced \shortname, a method to generate handheld objects conditioned on a 3D hand model or 2D hand image.
\shortnamethree operates entirely in 3D and takes as input a 3D hand model.
The key insight is to estimate \textbf{object codes} that capture the space of plausible objects and can be used to select plausible and diverse objects from a large dataset.
\shortnametwo takes this a step further by enabling us to edit hand images directly to either add or replace held objects.

\textbf{Limitations, Future Work, and Societal Impact}
Although our method works well and outperforms existing methods, it is still slow and can be improved for in-the-wild images -- we aim to address these limitations in future work.
As a method capable of image editing, our model can be used to misrepresent objects being held by people in images.
We plan to add watermarking capability to mitigate this problem.



\begin{ack}
This work was supported by NASA grant \#80NSSC23M0075, and NSF CAREER grant \#2143576.
\end{ack}

{
    \bibliographystyle{ieeenat_fullname}

}

\clearpage 
\appendix
\section{Additional Comparisons}
\label{sec:suppl_moreresults}

\begin{figure}[hp]
    \centering
    \includegraphics[width=1.0\textwidth]{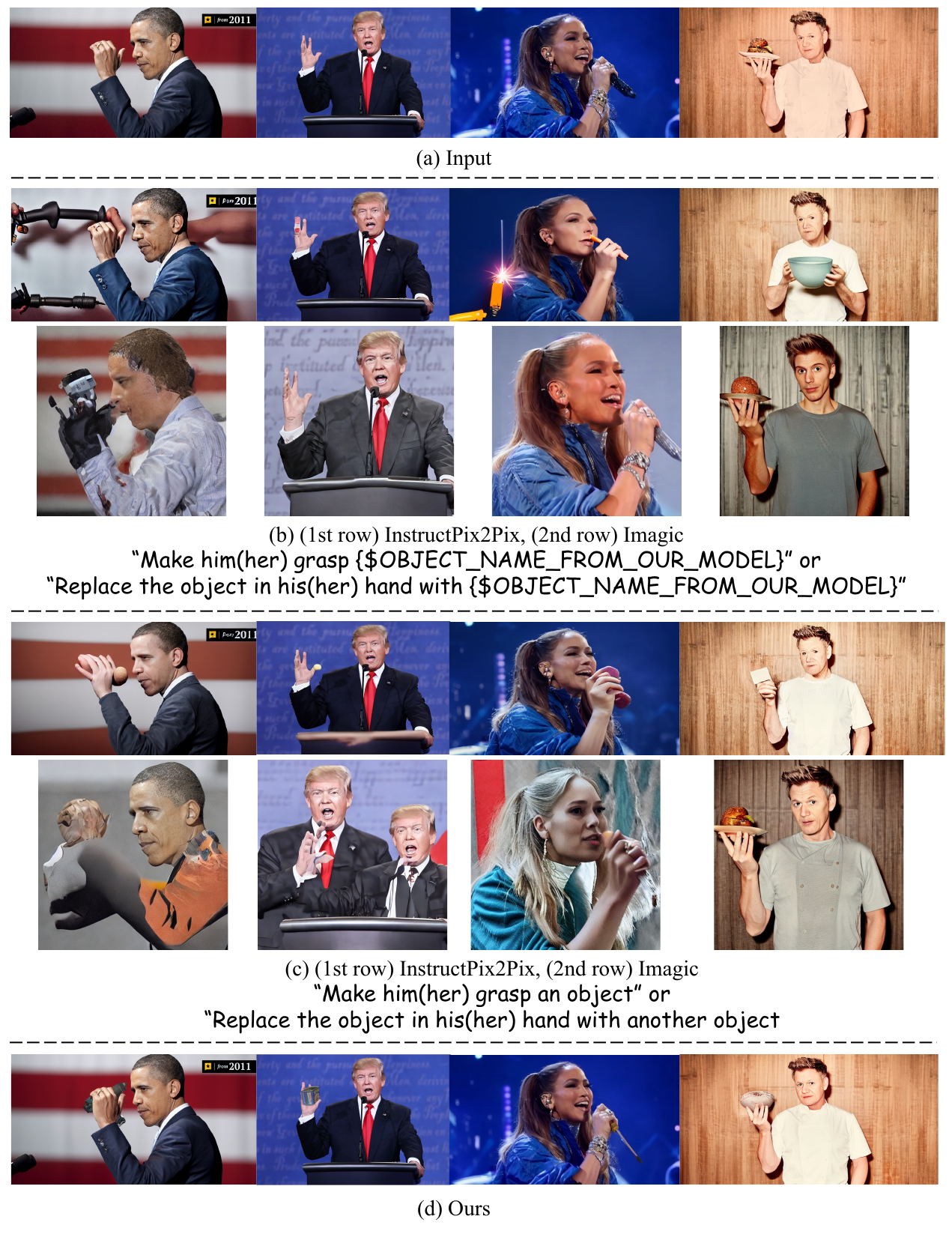}
    \caption{Additional comparisons of Fig.~\ref{fig:teaser}. Even with giving the object category as text prompts, the results are inferior to the results using the 3D grasping guide. Left to right: power drill, spam can, screwdriver, bowl.}
    \label{fig:famous_comparison}
\end{figure}

\begin{figure}[tp]
    \centering
    \includegraphics[width=0.9\textwidth]{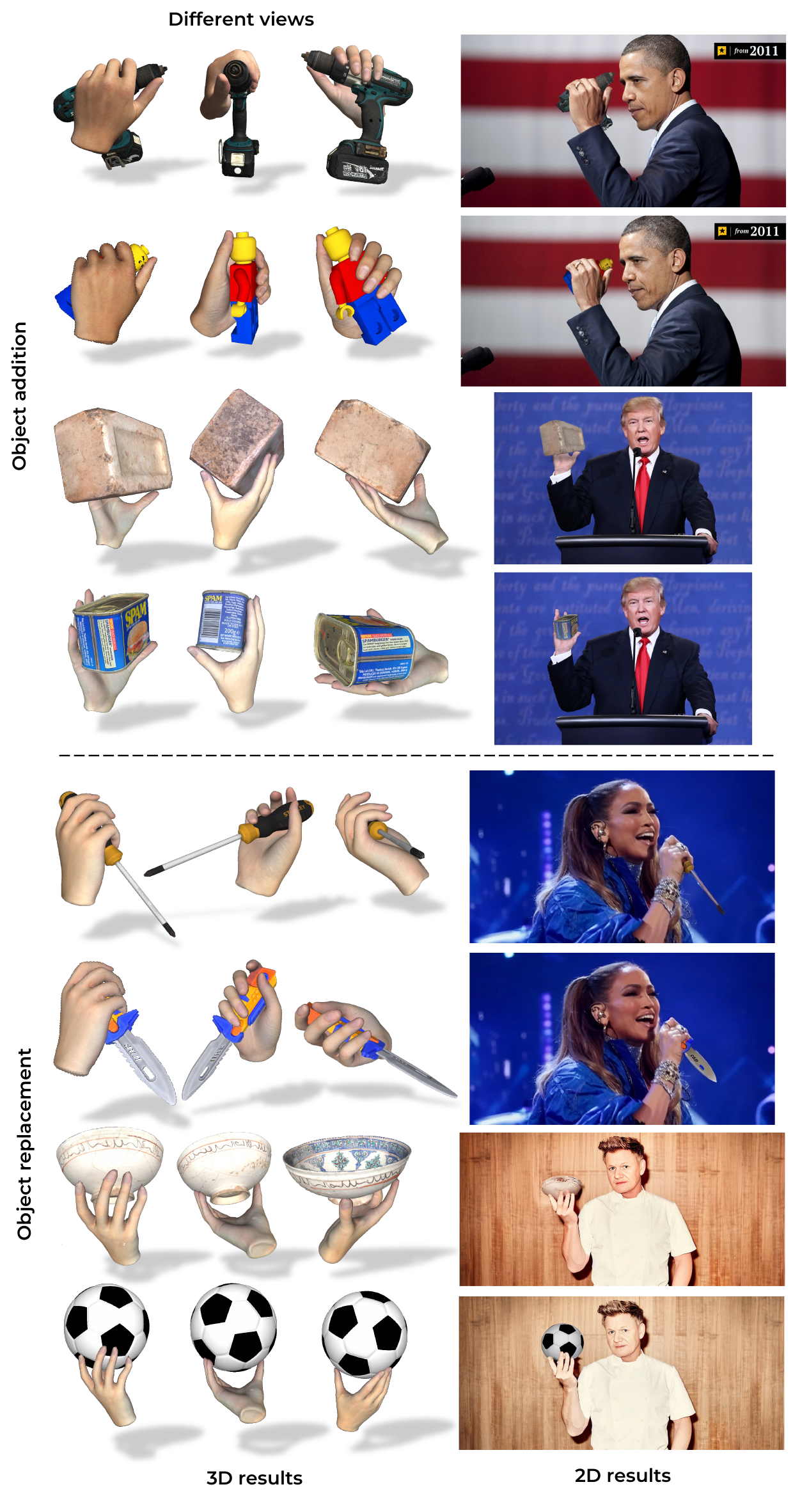}
    \caption{One hand pose can have multiple true answers. We sample different random noise $\bold{z}$ to get multiple answers.}
    \label{fig:extra_teaser}
\end{figure}

\begin{figure}[ht]
    \centering
    \includegraphics[width=1.0\textwidth]{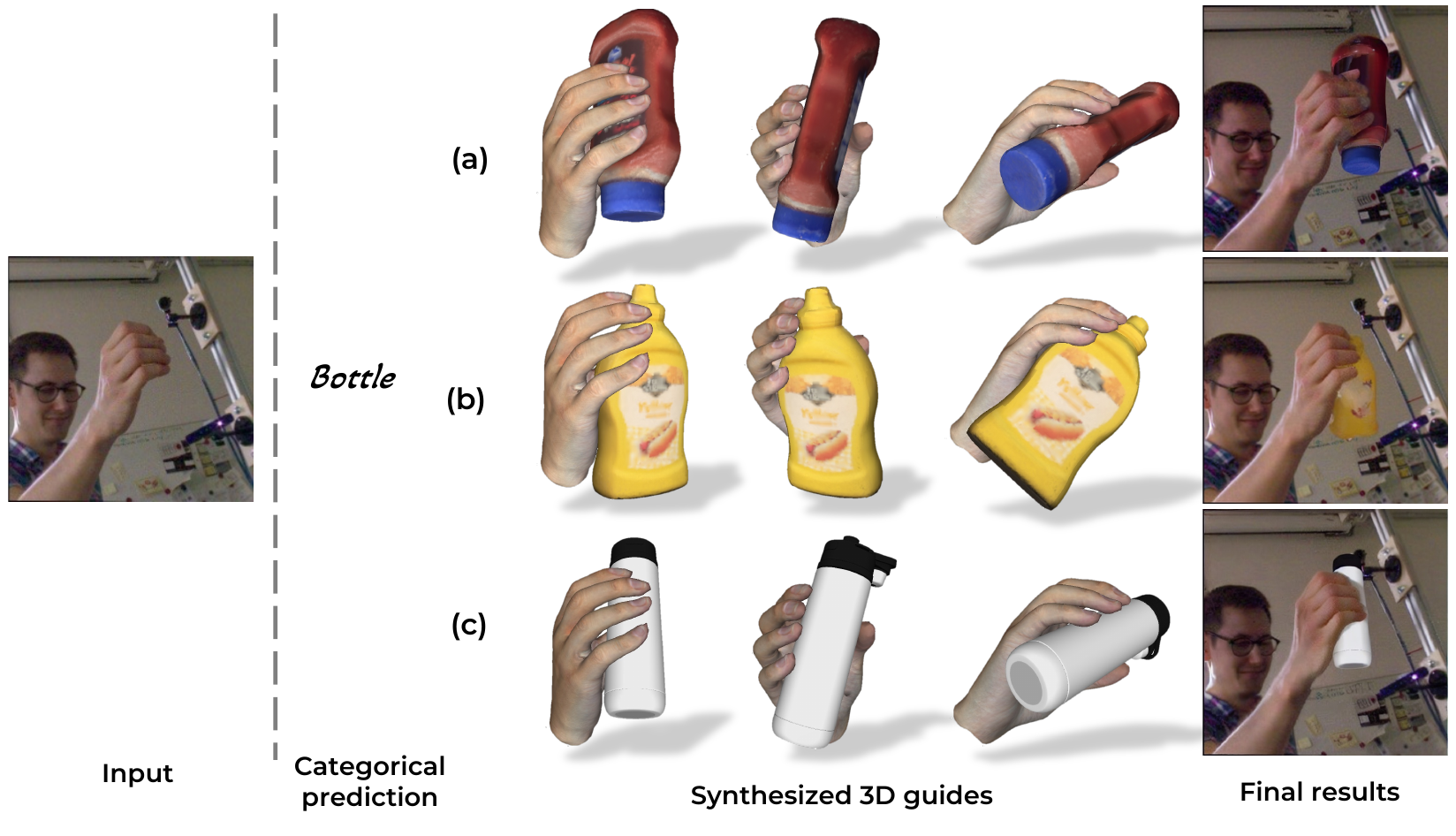}
    \caption{One categorical prediction produces different objects.}
    \label{fig:one_cate_many_objs}
    \vspace{-.5cm}
\end{figure}

\begin{figure}[tp]
    \centering
    \includegraphics[width=1.0\linewidth]{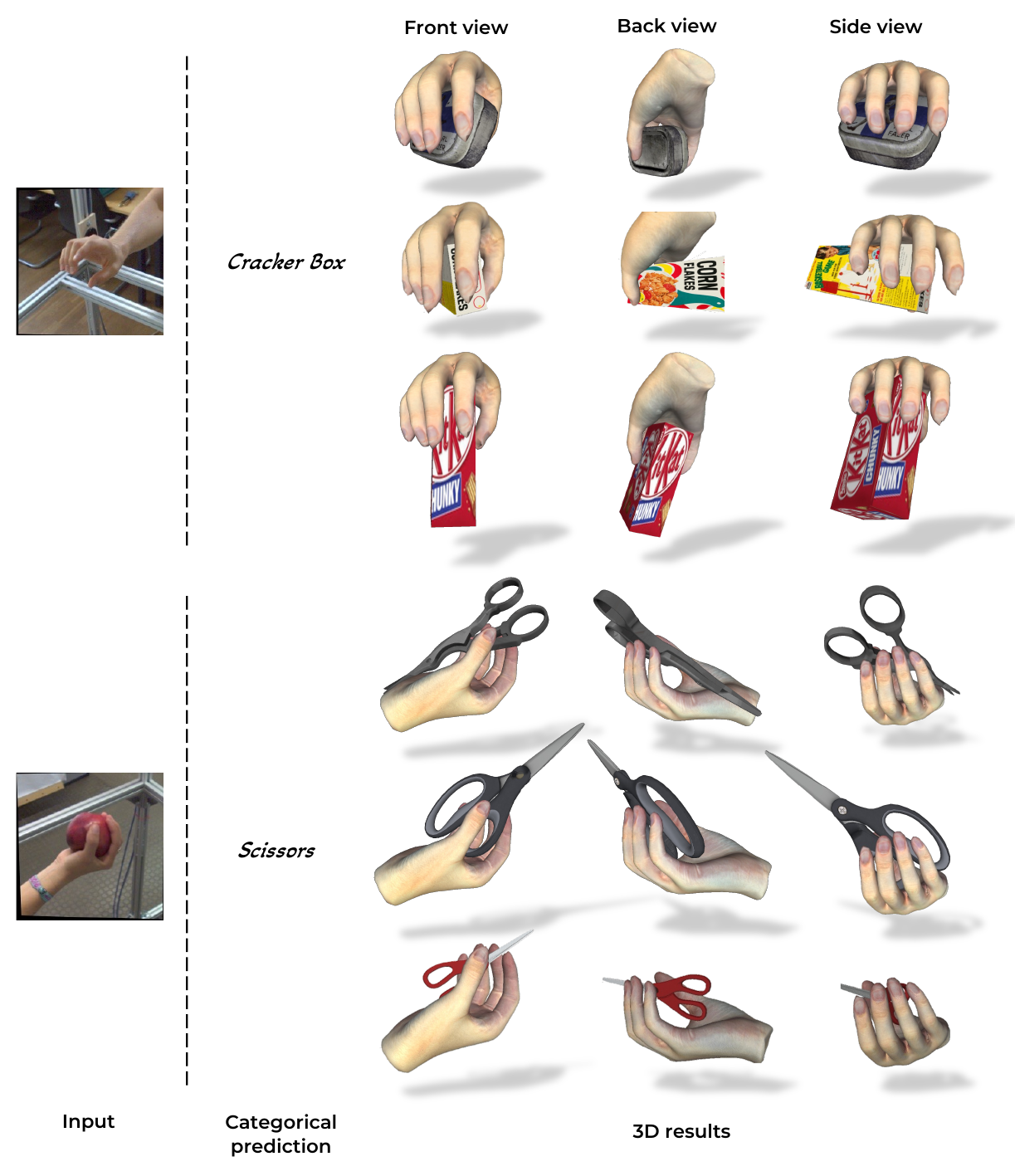}
    \caption{More 3D results. Our Object Shape code allows searching diverse plausible objects from Objaverse~\cite{objaverse} that matches the hand input.}
    \label{fig:moreresults_3D}
\end{figure}

\begin{figure}[tp]
    \centering
    \includegraphics[width=1.1\textwidth]{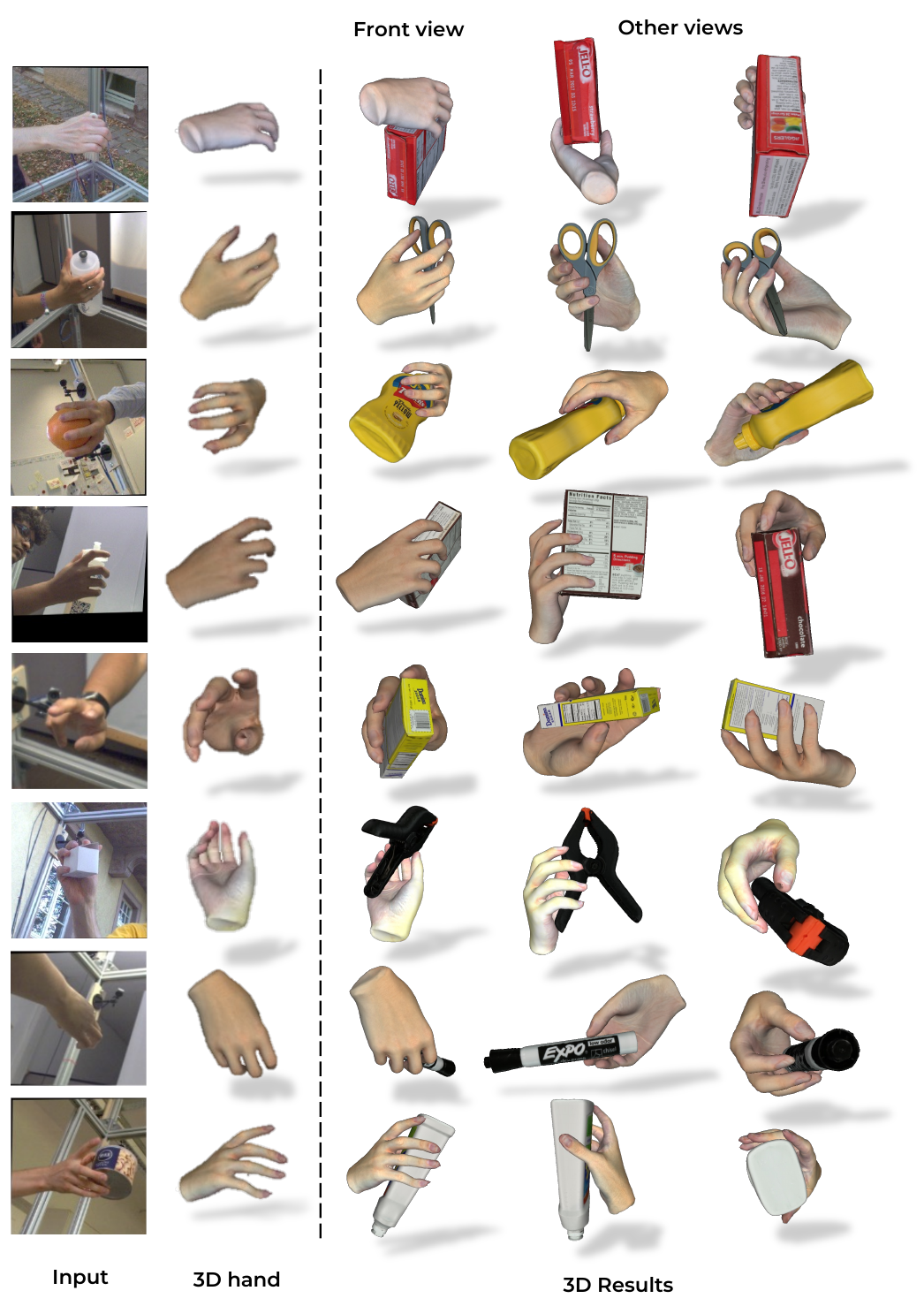}
    \caption{Our Object Shape code can also work with the YCB~\cite{ycb} objects.}
    \label{fig:moreresults_ycb}
\end{figure}

In fig.~\ref{fig:famous_comparison}, (b) For ~\cite{imagic, instructpix2pix}, even if we give our selected object to the text prompt, the object addition/replacement on hands fail, without our 3D guide. 
(c) Without knowing which object to grasp, the image editing models struggles to follow the prompt.

\section{Detailed Model Architecture}

\begin{figure}[ht]
    \centering
    \includegraphics[width=1.0\textwidth]{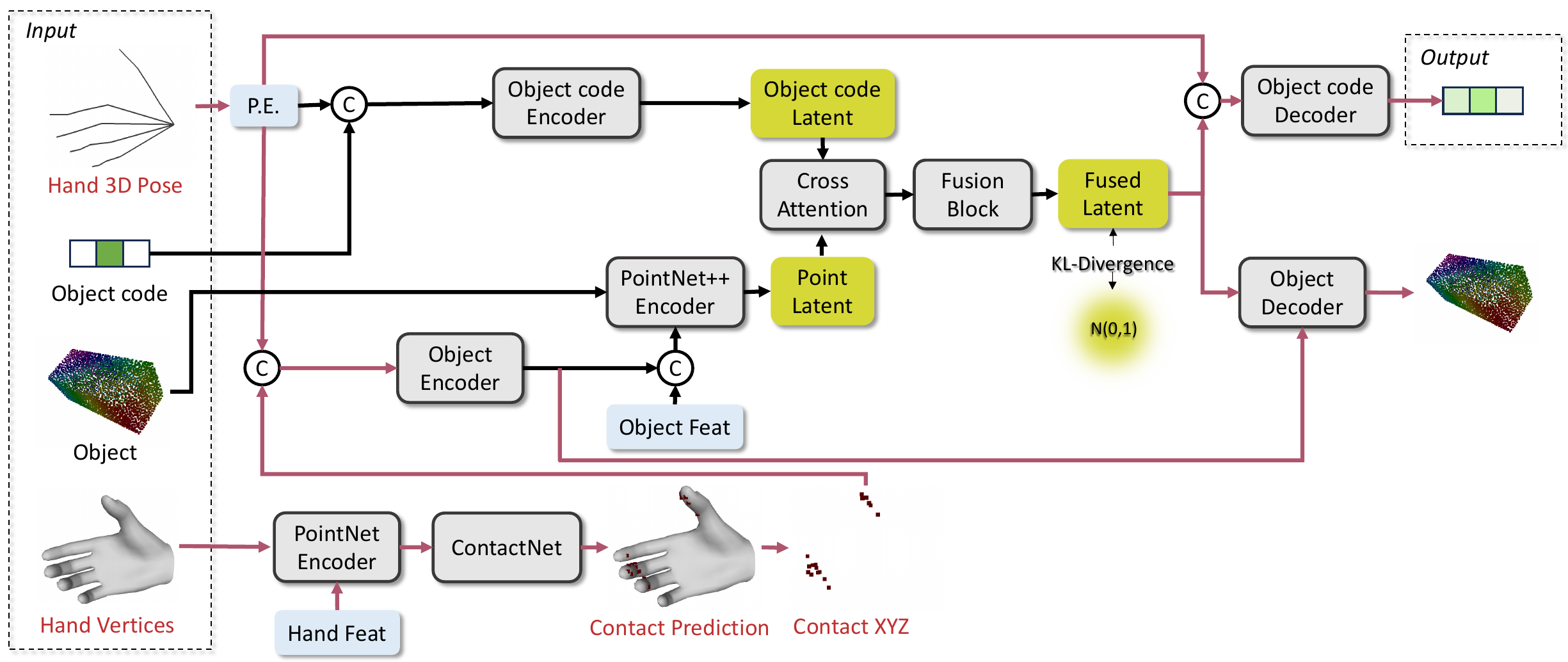}
    \caption{Details of the Object Selection network.}
    \label{fig:suppl_arch_detail}
    \vspace{-0.5cm}
\end{figure}

The red line shows the information that is available at inference.
P.E. means Positional Encoding~\cite{transformer, nerf} and the Object Feat and Hand Feat were obtained by the vertex normals. 
We incorporate PointNet~\cite{pointnet} to encode the hand vertices.
Unlike hand vertices, the number of object vertices varies depending on objects. 
Therefore, we leverage the global max pooling of PointNet++~\cite{pointnet++} to encode the object vertices.
For the hand mesh in the contact estimation, we use the canonicalized hand mesh, which is rotation normalized and centered around the world origin. 
Then, only the hand pose and shape remains for predicting the contact. 
By doing this, we make the prediction easier for the model. 
The ContactNet is inspired by GraspTTA~\cite{grasptta}.
However, it has differences in that our ContactNet only accepts the hand as a input. 
Following the GraspTTA, we use the architecture of CNNs with residual connections~\cite{resnet}. 
This ContactNet is separately trained in a supervised manner by DexYCB~\cite{dexycb}.
The predicted contact probability for hand vertices are converted into the xyz contact positions (Contact XYZ).
Then, the Contact XYZ and the Postional Encoded~\cite{nerf, transformer} are concatenated to the object encoder.
Together with the Object Feature, these information is fed into the PointNet++~\cite{pointnet++} encoder to make the Point Latent.
The Shape Code latent comes from the Shape Code encoder. 
The Shape Code encoder consists of CNN with residual connections~\cite{resnet}, following the human model model, SAGA~\cite{saga}.
The Shape Code latent and Point latent are entangled using the cross attention and the following Fusion Block of linear layers. 
Finally, the shape code is generated using the decoder.
The decoder is made up of the Shape Code decoder and the Object decoder, which follows the detail with ~\cite{saga}.
The total number of layers sums up to 102 by end-to-end.

\section{Non-Grasping Hand Rejection}
\label{sec:suppl_handrejection}

\begin{figure}
    \centering
    \includegraphics[width=0.8\linewidth]{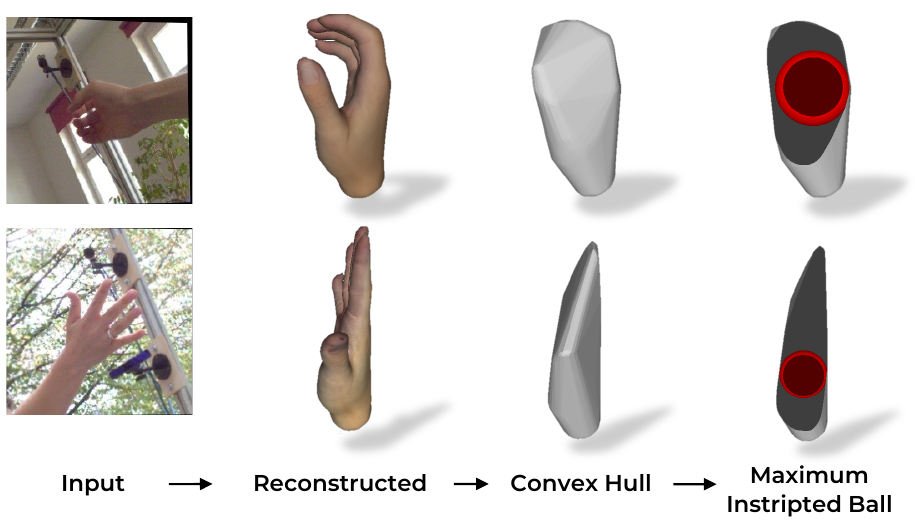}
    \caption{Non-grasping hand pose rejection algorithm (Sec.~\ref{sec:suppl_handrejection}). If the radius of maximum the inscripted ball is below a threshold, we reject that hand pose.}
    \label{fig:hand_rejection}
\end{figure}

In case the user gives an input image that does not contain the hand, the hand pose estiamtor and reconstructor can detect the existance of hands. 
However, without a non-grasping hand rejection algorithm, the model will automatically try to make an object to be grasped by that hand.
For example, we should reject a rest pose hand and some of the hand gestures, e.g., thumbs-up and making "two" with hands.
As a result, we propose a novel method to discriminate the grasping pose and the non-grasping pose. 
We illustrate this algorithm in Fig.~\ref{fig:hand_rejection}
Inversely using the grasp quality evaluation metrics from UGG~\cite{ugg} and DexGraspNet~\cite{dexgraspnet}, we leverage the convex hull of the reconstructed hand mesh and their maximum inscripted ball.
First, we compute the Convex Hull. 
Then, we obtain the inscripted ball, by minimizing a loss function, given by,
\begin{equation}
    \mathcal{L}_\text{ins\_ball} (o) = \text{SDF}(H, o) + \text{scale}(H),
\end{equation}
where the SDF takes the outside to be positive values. 
$o$ means the center of the inscripted ball, and the $H$ means the Convex Hull.
$\text{SDF}(X, x)$ means the signed distance function from a point $x$ to the mesh $X$.
We added the scale of the Convex Hull to avoid the loss function to be negative. 
Then, the radius of the inscripted ball follows, $\mathcal{L}_\text{ins\_ball}(o) - \text{scale}(H)$.
Even though the convex hull of a rest pose with the fingers all separate with each other (2nd row of Fig.~\ref{fig:hand_rejection} has similar size with the grasping pose (1st row of Fig.~\ref{fig:hand_rejection}, the radius of maximum inscripted ball of the Convex Hull has a strict difference.
This is because the maximum inscripted ball measures the thinnest part of the hand Convex Hull.
As a result, we threshold the radius of the maximum inscripted ball to discriminate the non-grasping hand.
If the radius is determined to be lower than 0.4 in the normalized hand space, we simply reject such input and ask for another input.

\section{Simulation Loss}
\label{sec:suppl_simloss}

\begin{figure}
    \centering
    \includegraphics[width=0.6\textwidth]{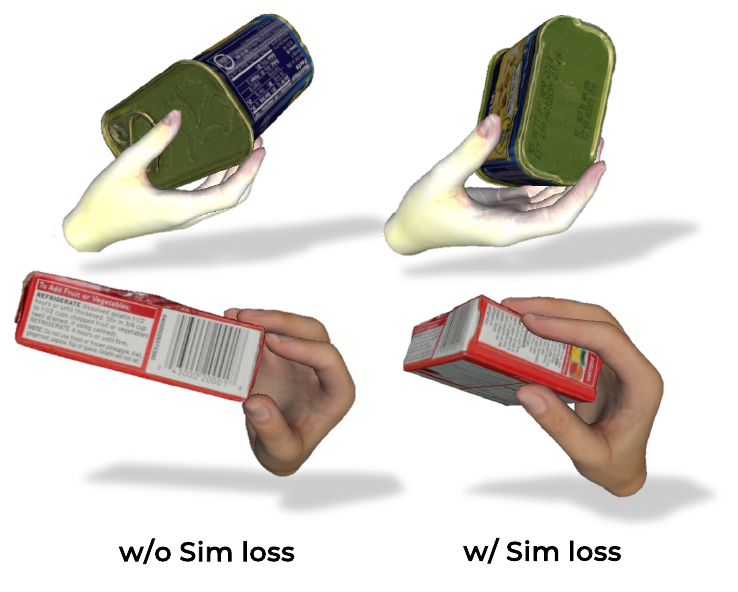}
    \caption{Ablation study of the Sim loss (Eq.~\ref{eq:sim_loss}).}
    \label{fig:force_closure}
\end{figure}

The ablation of the Simulation loss we incorporate in Sec. ~\ref{subsubsec:object_fitting} is illustraed in Fig.~\ref{fig:force_closure}. 
Note that our results on the left column still achieves almost perfect contacts and no penetration.
However, these graspings are imperfect, because the hands are not grasping the object with high physical stability.
Therefore, we add extra constraints to the Eq.~\ref{eq:fitting_loss}, i.e., Eq.~\ref{eq:sim_loss}.
Following ~\cite{forceclosure}, we ensure the force exerted to the object is "closed" by multiple direction, so that the object will stay still by the frictions of the hand and the object. 
Let us denote the contact points during optimization as $g_i, i=\left\{1, 2, .., N\right\}$, where N is the number of contact points.
For the friction cone axes, we compute the surface normals $\hat{n}_i$ from the contact vertices. 
Then, following the force closure~\cite{forceclosure_original} constraints,
\begin{equation}
    \mathcal{L}_\text{sim} = \text{eigen}_0 (GG' - \epsilon I) + ||G\hat{n}||_2 + \lambda_\text{dist} \sum_{g_i \in g} d(g_i, V_\text{obj})
    \label{eq:sim_loss}
\end{equation}
should be minimized, where
\begin{equation}
    G = \begin{bmatrix} I_{3 \times 3} & I_{3 \times 3} & ... & I_{3 \times 3} \\ \lfloor g_1\rfloor_\times & \lfloor g_2 \rfloor_\times & ...& \lfloor g_N \rfloor_\times\end{bmatrix}, 
    \label{eq:sim_G}
\end{equation}
and $\lfloor \cdot \rfloor_\times$ denotes skew-symmetric matrix.
$\text{eigen}_0$ is the smallest Eigenvalue. 

Below, we further justify the Eq.~\ref{eq:sim_loss} and the use of Eq.~\ref{eq:sim_G}.
When the $G$ matrix is given by Eq.~\ref{eq:sim_G} with the contact points $g_i, i=\{1, 2, ..., N \}$, ~\cite{forceclosure_original} introduced the constraints to satisfy the contact forces to make a closure as follows,
\begin{equation} \label{eq:forceclosure_derivation}
\begin{split}
    GG' \succcurlyeq \epsilon I_{6 \times 6},\\
    Gf = 0, \\
    f_i^T e_i > \frac{1}{\sqrt{\mu^2 + 1}}|f_i|, \\
    g_i \in S_i,
\end{split}
\end{equation}
where S is the object surface and the $f_i$ is the contact forces at the contact points $g_i$ and $e_i$ defines the friction cone axis. 
$\mu_i$ is the friction cone coefficient. 
$N$ denotes the number of contact points.
We find that the derivation from ~\cite{forceclosure} can effectively lead these constraints into Eq.~\ref{eq:sim_loss} by decomposing the $f$ into the normal component $f_n$ and the tangential component $f_t$.
Then, the second line of Eq.~\ref{eq:forceclosure_derivation} can be rewritten by,
\begin{equation}
\begin{split}
    Gf = f(f_n+f_t) = 0,\\
    G \frac{f_n}{||f_n||_2} = - \frac{Gf_t}{||f_n||_2},\\
    Gc = -\frac{Gf_t}{||f_n||_2},
\end{split}
\end{equation}
which automatically lead to the rewritten formulation of the second line of Eq.~\ref{eq:forceclosure_derivation} as follows,
\begin{equation}
    ||Gc||_2 < \delta
    \label{eq:forceclosure_Gc}
\end{equation}
, where $\delta$ is a small number. 
We exploit the vertex normal of the contact points  as the friction cone axis $e_i$.
In conclusion, following the first and last line of Eq.~\ref{eq:forceclosure_derivation} and the Eq.~\ref{eq:forceclosure_Gc}, we can easily reach the conclusion that the force closure is achieved by minimizing the loss presented in the Eq.~\ref{eq:sim_loss}.

\section{2D Image Editing}
\label{sec:suppl_2D}

In this section, we provide the missing details of Eq.~\ref{eq:score_based} and ~\ref{eq:diffeditor_interval}.
Following the DDIM~\cite{ddim} non-Markovian process, one can generate $x_{t-1}$ from $s_t$ as,
\begin{equation}
    x_{t-1} = \sqrt{\alpha_{t-1}} \left( \frac{x_t-\sqrt{1-\alpha_t} \epsilon_\theta^{(t)}(x_t)}{\sqrt{\alpha_t}} \right) + \sqrt{1-\alpha_{t-1}-\sigma^2_t} \cdot \epsilon^{(t)}_\theta (x_t) + \sigma_t \epsilon_t,
\label{eq:DDIM_step}
\end{equation}

where $\sigma_t = \eta \sqrt{(1-\alpha_{t-1})/(1-\alpha_t)} \sqrt{1-\alpha_t/\alpha_{t-1}}$.
The first term means the predicted $x_0$, and the second term indicates the direction pointing to $x_t$.
The last term is the random noise.
Affected by ~\cite{diffeditor}, we control the last term, i.e., the random noise, by introducing $\eta$ to the $\sigma_t$.
With $\eta=0$, Eq.~\ref{eq:DDIM_step} loses its randomness and becomes deterministic.
We use this for Eq.~\ref{eq:diffeditor_interval} to balance between the stochatic process and the deterministic process. 

Inspired by ~\cite{dragondiffusion}, the energy functions $\mathcal{E}$ in Eq.~\ref{eq:score_based} are given by,
\begin{equation}
    \mathcal{E}_\text{edit} = \frac{1}{\alpha + \frac{1}{2}\beta (1 + \text{cos}(f_i(\bold{F}^\text{gen}, \bold{m}^\text{gen}), \text{gc}(f_i(\bold{F}^\text{gud}, \bold{m}^\text{gud}))))}, f_i \in \left\{ f_\text{local}, f_\text{global}\right\},
\end{equation}
\begin{equation}
    \mathcal{E}_\text{content} = \frac{1}{\alpha + \frac{1}{2}\beta (1 + \text{cos}(f_\text{local}(\bold{F}^\text{gen}, \bold{m}^\text{share}), \text{gc}(f_\text{local}(\bold{F}^\text{gud}, \bold{m}^\text{share}))))},
\end{equation}
where $\bold{F}$ denotes the features and the masked features are $f_\text{local} = \bold{F}_t [\bold{m}]$ and $f_\text{global} = (\sum \bold{F}_t [\bold{m}])/\sum \bold{m}$.
The mask for the edited region, $\bold{m}_\text{edit}$ is aligned with the mask in the generated and reference images, which are denoted as $\bold{m}^\text{gen}$ and $\bold{m}^\text{gud}$, respectively. $\bold{m}^\text{share}$ denotes the shared region between the source and reference, e.g., background.
$\text{gc}(\cdot)$ is the gradient clipping operation~\cite{dragondiffusion}.
The energy function $\mathcal{E}_\text{edit}$ gets larger when the consine similarity between the masked features of the generated and reference gets larger.

\section{Design Choices}
\label{sec:suppl_designchoice}
\begin{figure}[ht]
    \centering
    \includegraphics[width=0.85\textwidth]{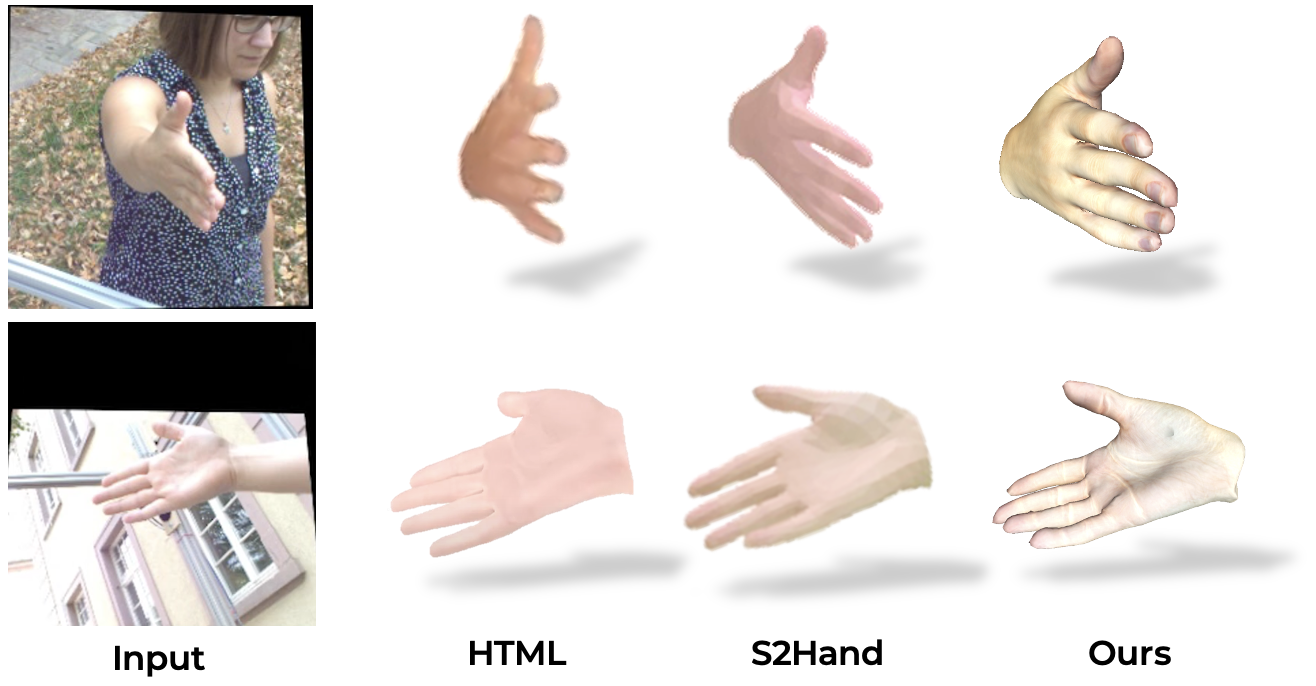}
    \caption{Comparisons of the single image hand textured reconstruction results. The superior quality of ours, which follows ~\cite{nimble} and ~\cite{hifihr}, motivated us to use it, other than the alternatives, i.e., HTML~\cite{html} and S$^2$Hand~\cite{s2hand}.}
    \label{fig:hand_recon_comparison}
\end{figure}
Our method performs "putting objects in people's hands" with a single image input only.
This simple input allows us to work with one of the most widely accessible input -- image. 
However, this brings an unique challenge of turning a single image into a high quality 3D representation.
There have been a few previous works which tried \emph{single image to textured 3D hand}.
HTML~\cite{html} introduced this task many year ago.
S$^2$Hand~\cite{s2hand} improved the training of such 3D hands with self-supervised learning. 
After that, NIMBLE~\cite{nimble} was introduced, which considered biometric bones and muscles and greatly increased the resolution of the hand mesh.
Then, some works~\cite{hifihr} tried to build a textured hand model using the NIMBLE. 
We found our high demand for a quality 3D avatar for hands to be a compelling motivation of applying the ~\cite{hifihr} to our hand reconstruction pipeline.
The comparison is depicted in Fig.~\ref{fig:hand_recon_comparison}

\section{Implementation Details}
\label{sec:imple_details}
We use 0.1 for the scaling range k.
We initialize the transformation of the object by overlapping the centers of the hand and object at the center and scaling the object by the predicted shape code. 
We used the pre-trained HAT~\cite{HAT} to perform Super Resolution on FreiHAND into 896 $\times$ 896.
For the 2D keypoints in~\ref{fig:2D_arch}, we depend on the Mediapipe~\cite{mediapipe}.
We used Stable Diffusion 1.5~\cite{stable_diffusion} as the backbone of the DiffEditor~\cite{diffeditor}.
We empirically selected $\eta_1 = 0.4$ and $\eta_2 = 0.2$.
When making the $\bold{h}_c$, we threshold the distance between hand and object by 1cm.
We used ~\cite{hifihr}, which is based on the biometric hand parameterized model~\cite{nimble}, for acquiring the hand 3D model from a single image. 
The reason for the design choice of this hand 3D model is described in Sec.~\ref{sec:suppl_designchoice}.

\section{Computational Resources}
\label{sec:compute_resources}
The Object Selection Network was trained for 12 hours on two RTX 3090 gpus. 
The inference of our model consists of the object selection time, the Objaverse fetching time, the optimization time, and the final editing time. 
Apart from the editing, it took us, in average, 56s for an input on one RTX 3090 gpu.
The time can vary depending on the Internet speed, and we tested our model with a network of around 865 Mbps. 
The overall inference time comparisons are listed in Tab.~\ref{tab:2D_image}. 
Although our method adds extra time for object selection and 3D grasping, the large improvement upon other methods on the results could justify the cost.

\section{Limitations}
{\bf Limitations} 
The performance of our method depends on the effectiveness of the off-the-shelf single image textured reconstruction model and the 2D keypoint estimation model. 
Moreover, our current method does not support open-vocabulary object categories, nor infinite number of objects. 
We believe that although the use of large-scale 3D models provided us with objects with high details, the growing community of shape generation~\cite{clip_sculptor, meshgpt} and texture synthesis~\cite{mapa, paintit} methods would be able to substitute such large, but predefined 3D database in our future works. 
Plus, incorporating human affordance prior into our test-time optimization will be beneficial, given the rising popularity of large pretrained models that may better understand human behaviors. 

\clearpage

\end{document}